\PassOptionsToPackage{table}{xcolor}
\documentclass[11pt, a4paper, copyright]{google}

\usepackage[authoryear, sort&compress, round]{natbib}
\makeatletter

\usepackage[utf8]{inputenc} % allow utf-8 input
\usepackage[T1]{fontenc}    % use 8-bit T1 fonts
\usepackage[misc]{ifsym}
\usepackage{wrapfig}
\usepackage[colorlinks=true, linkcolor=blue, citecolor=blue, urlcolor=blue]{hyperref}
\usepackage{xurl}
\usepackage{url}
\usepackage{booktabs}
\usepackage{graphicx}
\usepackage{multirow}
\usepackage{dsfont}
\usepackage{arydshln}
\usepackage{enumitem}
\usepackage{amsmath}
\usepackage{amssymb}
\usepackage{caption}
\usepackage{subcaption}
\usepackage{tabularx}
\usepackage{colortbl}
\usepackage{tikz}
\usetikzlibrary{positioning, shapes.geometric}
\usepackage{fvextra}
\usepackage{dashrule}
\usepackage{algorithm}
\usepackage{algpseudocode}
\usepackage{comment}
\usepackage{nicefrac}
\usepackage{microtype}
\usepackage{array}
\usepackage{makecell}
\usepackage{newunicodechar}
\usepackage{pifont}
\usepackage{xspace}
\usepackage[most]{tcolorbox}

\newunicodechar{✓}{\ding{51}}
\newunicodechar{✗}{\ding{55}}

\newtcolorbox{casebox}[2]{breakable, enhanced, colback=#2, colframe=#1,
  boxrule=1pt, arc=6pt, left=5pt, right=5pt, top=4pt, bottom=4pt,
  before skip=10pt, after skip=12pt}
\newenvironment{centerpage}{\vspace*{\fill}}{\par\vspace*{\fill}}

\newcommand{\method}{\textsc{DataOrchestra}\xspace}

\definecolor{boxfill}{rgb}{0.95,0.97,1.0}
\definecolor{boxstroke}{rgb}{0.2,0.4,0.7}
\definecolor{panelblue}{rgb}{0.86,0.95,0.99}
\definecolor{grpA}{RGB}{255,235,220}
\definecolor{grpB}{RGB}{235,255,235}
\definecolor{grpC}{RGB}{250,235,240}
\definecolor{grpD}{RGB}{255,250,228}
\definecolor{grpE}{RGB}{228,248,246}
\definecolor{grpF}{RGB}{244,234,252}
\definecolor{chunkcolor}{rgb}{0.78,0.15,0.12}
\definecolor{goodfill}{RGB}{230,245,232}
\definecolor{goodstroke}{RGB}{70,150,85}
\definecolor{badfill}{RGB}{252,231,236}
\definecolor{badstroke}{RGB}{198,80,105}

\newcommand{\promptfont}{\fontsize{7pt}{7.8pt}\selectfont}
\newcommand{\promptfontsmall}{\fontsize{6pt}{6.1pt}\selectfont}

\makeatother

\uselogo{}

\title{DataOrchestra: Learning to Orchestrate Per-Example Curation of Pretraining Data}

\author[1,3,4]{Zhen Huang}
\author[1,3,4]{Yikun Wang}
\author[2,3,4]{Shijie Xia}
\author[2,3,4 \Letter]{Pengfei Liu}

\affil[ \Letter]{Corresponding author}
\affil[1]{Fudan University}
\affil[2]{Shanghai Jiao Tong University}
\affil[3]{SII}
\affil[4]{GAIR}

\begin{abstract}

  \vspace{-0.2in}

  {\fontsize{11pt}{11pt} \selectfont \raisebox{-0.06em}{\includegraphics[height=0.9em]{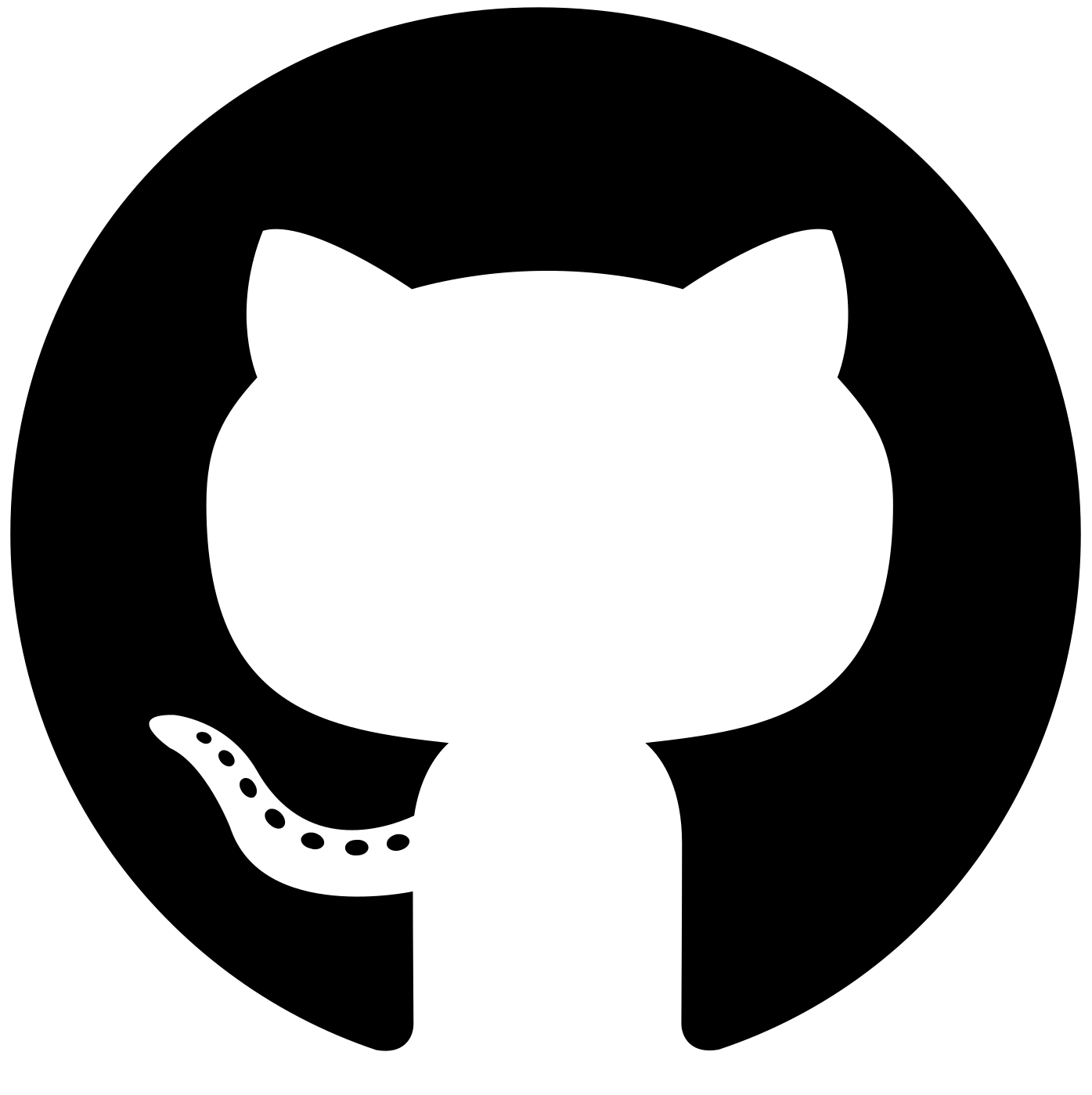}} Code: \href{https://github.com/GAIR-NLP/DataOrchestra}{https://github.com/GAIR-NLP/DataOrchestra}  }
  \\

   Pretraining data processing is critical to the downstream performance of Large Language Models (LLMs). However, many existing approaches define a fixed processing strategy at the corpus or domain level and apply it uniformly to many examples, without adapting to the needs of each example. We propose \method, a framework that unifies different processing operations and orchestrates an example-specific pipeline for each example. Given a chunk of pretraining data, an orchestrator decides whether to drop, untouch, or clean it. For a chunk to be cleaned, it selects one or more downstream operations, ranging from programmatic editing to different forms of LLM-based rewriting. For each rewriting step, it further generates a concrete instruction, which is executed by the corresponding downstream tool model. We pretrain models from 0.5B to 7B from scratch on web data processed by \method and observe stable average gains over individual data-processing methods across 11 benchmarks. \method is also effective for math continued pretraining and outperforms stronger processing baselines, while reducing processing compute by skipping unnecessary downstream operations.
\end{abstract}

\begin{document}

\maketitle

\section{Introduction}

Large Language Models (LLMs) are increasingly used to process pretraining data because of their flexibility in handling diverse text. Some methods score and filter entire low-quality documents~\citep{wettig2024qurating,peng2025dataman,yu2024mates,engstrom2024dsdm}, while others perform finer-grained processing within documents, which we group into three types. At the lightest level, LLMs generate programs that edit line-level noise, such as ads and boilerplate, without rewriting the entire text; we call this \textit{noise pruning (NP)}~\citep{zhou2024programming,bi2025refinex}. With stronger intervention, LLMs rewrite the entire text to repair formatting, grammar, tables, or other surface-level issues while preserving the original content, which we term \textit{surface rectification (SR)}~\citep{maini2024rephrasing,nguyen2025recycling,yu2025repro}. At the strongest level, LLMs rewrite knowledge- or reasoning-intensive data, such as math or Wikipedia text, to add explanations or improve its educational value, which we term \textit{pedagogical augmentation (PA)}~\citep{team2025kimi,fujii2025rewriting,qin2026data}. Together, these methods cover a wide range of data-processing needs, but no existing framework integrates them and adaptively decides, for each example, which operations to use and how each should be applied.

\begin{figure}[t]
\centering
\includegraphics[width=\linewidth]{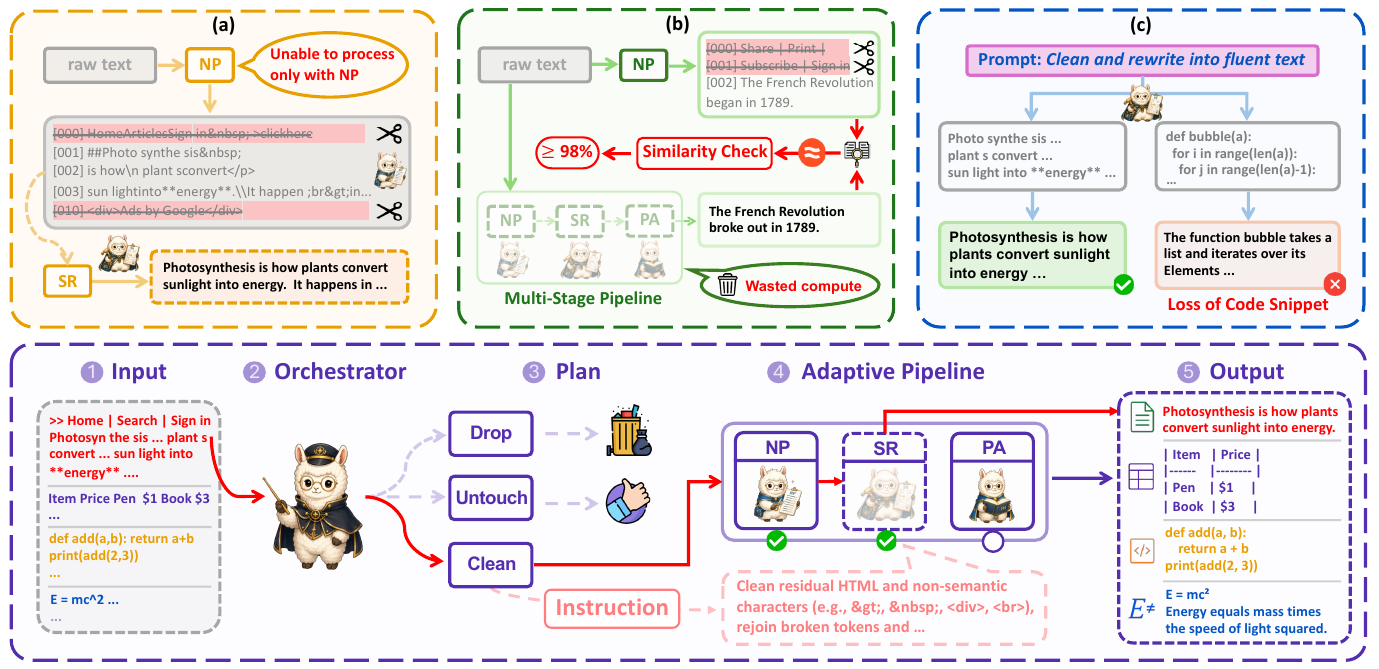}
\caption{\textbf{Top:} limitations of existing pretraining-data curation that motivate our design. \textbf{Bottom:} the overview of our \method framework.}
\label{fig:overview}
\end{figure}

Specifically, existing methods have three main limitations. First, many methods focus on a single data-processing operation, but a single operation cannot address all types of data issues. For example, programmatic noise pruning methods like ProX~\citep{zhou2024programming} can remove line-level noise but cannot repair more complex corruption that requires rewriting (Figure~\ref{fig:overview}a). Second, even methods that combine multiple operations often apply the same multi-stage pipeline to diverse examples. However, different examples require different treatments: some low-quality examples should be dropped rather than repaired~\citep{niklaus2026can,maini2025beyondweb}, some high-quality ones should remain untouched to avoid over-deletion~\citep{bi2025refinex} or hallucination~\citep{huang2025survey}, and others may require only a subset of processing stages. Applying the full pipeline to every example can therefore waste compute with little additional benefit (Figure~\ref{fig:overview}b). Third, when LLM-based rewriting is needed, most methods use a shared prompt across diverse examples. Yet different examples may require different rewriting goals, so one general prompt cannot fit all cases (Figure~\ref{fig:overview}c). Recent work explores different rewriting strategies across different domains~\citep{mi2026data}, but such adaptation remains coarse-grained. In short, these limitations motivate a framework that can unify different processing operations, select an example-specific subset of them, and further adapt each rewriting step to the content being processed.

To fill this gap, we propose \method, a framework that orchestrates an example-specific processing pipeline for each data example, as shown in the bottom of Figure~\ref{fig:overview}. Given a chunk, the orchestrator first decides whether to drop, untouch, or clean it. For a chunk to be cleaned, it selects which processing stages to apply and orchestrates them into an adaptive pipeline. For each stage involving LLM rewriting, the orchestrator further generates a concrete, example-specific instruction (e.g., which table to fix, which explanation to add), which is passed to the downstream tool model that actually performs the operation. To train the orchestrator, we first use a teacher LLM to propose an initial processing plan for each chunk. We then execute this plan with downstream tool models and verify the resulting changes. Based on the execution feedback, we evolve the initial plans by removing stages that produce only minimal changes or harm the original content, and by refining rewriting instructions when the output loses information, introduces factual errors, or still contains surface-level noise. This evolution grounds the plans in the actual behavior of downstream tool models rather than relying only on the teacher LLM's high-level judgment. In this way, we construct 300K pairs of high-quality training samples and use them to fine-tune the orchestrator.

To verify its effectiveness, we apply \method to four common web datasets, RedPajama-V2~\citep{weber2024redpajama}, DCLM-RefinedWeb~\citep{li2024datacomp}, C4~\citep{raffel2020exploring}, and FineWeb~\citep{penedo2024fineweb}, instantiated with tool models no larger than 4B: a 0.6B noise pruning model and a 4B rewrite model (Qwen3-4B). Pretraining models from scratch from 0.5B to 7B, we observe stable average gains over other single data-processing methods across 11 benchmarks. The same framework also works on math continued pretraining setting, improving the quality of OpenWebMath~\citep{paster2024openwebmath} and MegaMath~\citep{zhou2025megamath} and yielding gains on scientific reasoning benchmarks. \method further outperforms stronger baselines, including rewriting methods that filter and mix rewritten data with the original corpus using fastText, as well as several fixed multi-stage processing pipelines, while reducing overall processing compute.

In summary, our contributions are as follows:

\begin{itemize}[leftmargin=*]
   \item We propose \method, a framework that orchestrates an example-specific processing pipeline for pretraining data.
   \item We validate its effectiveness across multiple datasets, settings, and model sizes, and show that it improves data quality while saving compute by skipping unnecessary stages.
   \item We will release the orchestrator model, together with the full scripts and code of the framework, to support further research by the community.
\end{itemize}

% In summary, our contributions are as follows:
% (1) We propose \method, a plug-in framework that orchestrates an example-specific processing pipeline for pretraining data.
% (2) We validate its effectiveness across multiple datasets, settings, and model sizes, and show that it improves data quality while saving compute by skipping unnecessary stages.
% (3) We will release the orchestrator model, together with the full scripts and code of the framework, to support further research.

\section{Related Work}

\paragraph{Pretraining Data Selection and Processing}
As pretraining scales to the trillion-token level, the community increasingly focuses on data quality over quantity~\citep{gunasekar2023textbooks}, since web-crawled data (e.g., CommonCrawl) is noisy and training on it directly wastes compute or hurts performance. Early work filters low-quality documents along dimensions such as language~\citep{wenzek2020ccnet}, web URLs~\citep{penedo2023refinedweb,penedo2024fineweb}, and heuristic rules like length and character counts~\citep{rae2021scaling,raffel2020exploring}, and removes duplicates through deduplication~\citep{lee2022deduplicating,abbas2023semdedup}. To improve selection, model-based approaches either train a binary fastText classifier~\citep{joulin2017bag,li2024datacomp} or a small model that scores quality along multiple dimensions~\citep{wettig2024qurating,yu2024mates,peng2025dataman}, though some work questions the necessity of filtering altogether~\citep{mohri2026bitter}. Beyond document-level filtering, another line performs in-document refinement, using heuristic rules~\citep{rae2021scaling,penedo2023refinedweb} or a small LLM that edits text through programs~\citep{zhou2024programming,bi2025refinex}.

\paragraph{Synthetic Pretraining Data Curation}
Using LLMs to clean and synthesize data has become common practice~\citep{su2025nemotron}. Rules or programs from lightweight models cannot fix complex issues such as broken tables, formulas, or grammar, whereas rewriting with LLMs can recycle such data: WRAP~\citep{maini2024rephrasing} and ReWire~\citep{nguyen2025recycling} rephrase data into specific formats, while RePro~\citep{yu2025repro} trains a faithful rewriter with reinforcement learning. Other work synthesizes diverse data through role-playing~\citep{hao2025reformulation}, knowledge-graph synthesis~\citep{yang2024synthetic}, and direct continuation~\citep{benallal2024cosmopedia}, and for knowledge- or reasoning-intensive text, enhancing its educational value improves utility~\citep{qin2026data,team2025kimi,fujii2025rewriting}. However, synthetic data also risks model collapse~\citep{gerstgrasser2024model} or hallucinated and factual errors~\citep{liu2024best}, and how to use it effectively remains open~\citep{maini2025beyondweb,niklaus2026can}. Overall, existing methods either treat each operation in isolation or apply the same multi-stage pipeline to large amounts of data; none combines operations of different levels into a pipeline tailored to each example.

\section{Methods}

\subsection{Task Formulation}
We treat pretraining data curation as a transformation over text applied at the level of chunks. Given a document $D$ from a corpus $\mathcal{C}$, we segment it into an ordered sequence of chunks $\mathrm{Split}(D) = (c_1, \dots, c_n)$, where each chunk holds at most $W$ tokens (we use $W=1024$\footnote{All token counts in this paper use the Qwen3 tokenizer.}). We process each chunk $c_i$ with its own pipeline $P_i = s_1 \to \cdots \to s_k$, an ordered composition of one or more stages. Formally, each stage $s$ maps a chunk to a new chunk and may return the empty string $\varnothing$ to delete it; an empty pipeline thus leaves a chunk unchanged, while a pipeline whose output is $\varnothing$ removes it. We then rebuild the refined document $\hat{D} = \hat{c}_1 \,\Vert\, \cdots \,\Vert\, \hat{c}_n$ by concatenating the processed chunks $\hat{c}_i = P_i(c_i)$ in their original order and dropping any that were deleted.

We organize data processing into three high-level decisions: \emph{Drop}, \emph{Untouch}, and \emph{Clean}. \emph{Drop} removes a chunk with no usable content~\citep{wettig2024qurating,yu2024mates,peng2025dataman}, while \emph{Untouch} keeps it unchanged when no processing is needed. \emph{Clean} applies one or more processing stages. We group cleaning operations into three stages with increasing levels of intervention. \emph{Noise Pruning (NP)} uses a lightweight small LLM to generate programmatic edits that remove clearly worthless lines, such as boilerplate, ads, links, and navigation bars~\citep{zhou2024programming,bi2025refinex}. \emph{Surface Rectification (SR)} uses an LLM to repair issues that line-level edits cannot fix, such as broken tables and formulas, grammatical errors, and disordered layout, while preserving the original meaning~\citep{maini2024rephrasing,yu2025repro}. \emph{Pedagogical Augmentation (PA)} rewrites knowledge- or reasoning-intensive text to improve its educational value by expanding existing points, adding relevant knowledge, and making the underlying reasoning more explicit~\citep{nguyen2025recycling,qin2026data,fujii2025rewriting}. A clean pipeline may contain any subset of these stages, applied in the order NP $\rightarrow$ SR $\rightarrow$ PA.

\begin{table}[t]
\caption{The operations in \method, organized by the orchestrator's two-level decision. At the top level, a chunk is assigned to \emph{drop}, \emph{untouch}, or \emph{clean}; a clean chunk further passes through one or more of the three cleaning stages NP, SR, and PA, applied in this order. The last column lists representative prior work for each operation.}
\label{tab:operations}
\begin{center}
\small
\begin{tabularx}{\linewidth}{@{}l l l X@{}}
\toprule
\textbf{Decision} & \textbf{Operation} & \textbf{Operator} & \textbf{Relevant Methods} \\
\midrule
Drop    & ---  & \texttt{drop(c)} $\rightarrow$ $\varnothing$ & QuRating, MATES, DataMan \\
\midrule
Untouch & ---  & \texttt{untouch(c)} $\rightarrow$ \texttt{c}  & --- \\
\midrule
\multirow{3}{*}{Clean}
        & NP & \texttt{np(c)} $\rightarrow$ \texttt{remove\_lines(start, end)} & ProX, RefineX \\
        & SR & \texttt{sr(c)} $\rightarrow$ \texttt{rewrite(instructions)}     & WRAP, RePro \\
        & PA & \texttt{pa(c)} $\rightarrow$ \texttt{rewrite(instructions)}     & ReWire, Darwin-Science \\
\bottomrule
\end{tabularx}
\end{center}
\end{table}

\subsection{\method Framework}
\label{sec:framework}

\method centers on the orchestrator model, which decides how each chunk is processed and assigns the actual execution to a set of tool models. Given a chunk $c$, the orchestrator first makes a high-level decision among \emph{drop}, \emph{untouch}, and \emph{clean} (Table~\ref{tab:operations}). A drop decision removes the chunk and an untouch decision keeps it unchanged, ending its pipeline immediately. If the decision is clean, the orchestrator selects which of the three stages, NP, SR, and PA, the chunk passes through, and the chunk flows through the selected stages in order, each handled by its own tool model. For NP, the noise pruning tool model returns one or more \texttt{remove\_lines(start, end)} operations that delete lines \texttt{start} to \texttt{end}; unlike ProX and RefineX, we keep only whole-line removal and drop in-line substring edits, which simplifies the duty of this small tool model. For SR and PA, the chunk is passed to the corresponding rewriting tool model with a two-part instruction: a general part, shared by all chunks routed to the stage, that states the basic rewriting principles, and a chunk-specific part that the orchestrator generates from the chunk, specifying what it needs, such as which formatting to repair for SR or which explanation to add for PA.

\subsection{Orchestrator Model Training}
\label{sec:orchestrator_model_training}

\begin{figure}[t]
\centering
\includegraphics[width=\linewidth]{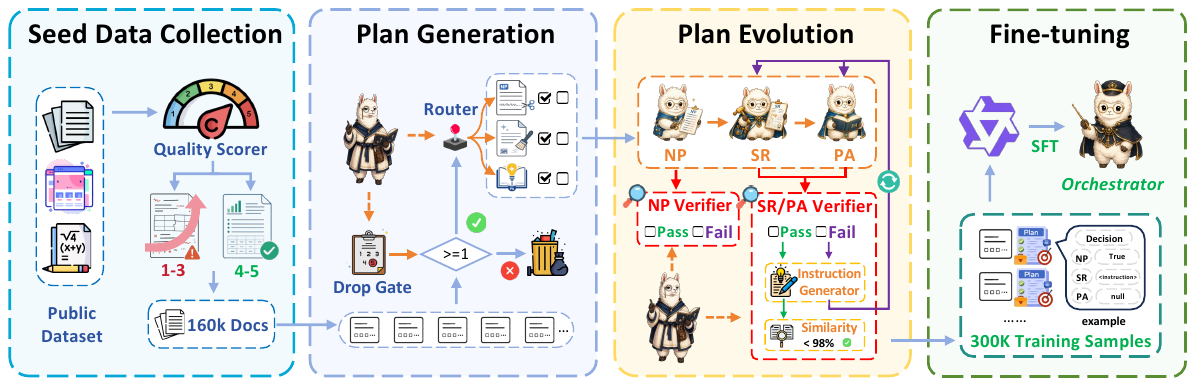}
\caption{The overview of our orchestrator model training pipeline.}
\label{fig:orchestrator_training}
\end{figure}

Figure~\ref{fig:orchestrator_training} gives an overview of how we train the orchestrator. Starting from a pool of seed documents, we generate an initial coarse-grained plan for each chunk and evolve it into a fine-grained plan through actual tool-model execution and verification, yielding a chunk-specific instruction for every rewriting stage. We then fine-tune the orchestrator on the resulting data.

\paragraph{Seed Data Collection}
We build the seed data for training the orchestrator from widely used public pretraining corpora: RedPajama-V2, DCLM-RefinedWeb, C4, and FineWeb for the general domain, and OpenWebMath and MegaMath for the science domain. We reserve a held-out portion of each corpus for orchestrator training and keep the rest for the pretraining experiments in Section~\ref{sec:exp}; collection and sampling details are in Appendix~\ref{app:data_collection}. Since a seed set dominated by clean documents would leave most chunks untouched and give little training signal, we score every document with DataMan~\citep{peng2025dataman}, an LLM-based quality scorer that assigns a score from $1$ to $5$, and upsample low-quality documents so that those scored $1$--$3$ and $4$--$5$ form two equal-sized groups. This gives a final pool of $160$K documents for orchestrator training.

\paragraph{Coarse-grained Plan Generation}
% For every chunk of the documents collected above, we use a teacher LLM\footnote{All teacher LLMs and verifier LLMs used in this paper, including the one for retraining the ProX model, are Qwen3-235B-A22B-Instruct-2507.} to produce a coarse-grained plan in two steps. First, we decide whether to drop the chunk. Unlike prior methods that drop based on educational value or format score, we lower the threshold and discard a chunk only when it has no salvageable value, giving each chunk a better chance of being repaired; concretely, we prompt the teacher LLM to rate the chunk from $0$ to $5$ under a stricter rubric and drop it only when the score is below $2$. Second, for each chunk that is not dropped, we ask the same teacher LLM for a binary True/False judgment on each of NP, SR, and PA, indicating whether the chunk needs that stage. A chunk that survives the drop step but receives all-False judgments is marked untouch. Together, the drop decision, the untouch decision, and the three stage judgments form the coarse-grained plan for each chunk. Prompts used for this procedure are given in Appendix~\ref{app:prompt_coarse_plan}.

For chunks from seed data, we use a teacher LLM\footnote{All teacher LLMs and verifier LLMs (introduced later) used in this paper, including the one for retraining the ProX model, are Qwen3-235B-A22B-Instruct-2507.} to produce a coarse-grained plan in two steps. First, we decide whether to drop the chunk. Unlike prior methods that drop based on educational value or format score, we lower the threshold and discard a chunk only when it has no salvageable value, giving each chunk a better chance of being repaired: we prompt the teacher LLM to rate the chunk from $0$ to $5$ under a stricter rubric and drop it only when the score is below $2$. Second, for each surviving chunk, we ask the same teacher LLM for a binary judgment on each of NP, SR, and PA, indicating whether the chunk needs that stage; a chunk that receives all-False judgments is marked untouch. The drop decision, the untouch decision, and the three stage judgments together form the coarse-grained plan. Prompts are given in Appendix~\ref{app:prompt_coarse_plan}.

\paragraph{Fine-grained Plan Evolution}

A coarse plan still has two limitations. First, the teacher's high-level decisions may not match what the tool models actually do: a stage marked as necessary may change the chunk only marginally, or even damage it, for example by over-pruning non-noise lines or hallucinating during rewriting. Second, for the rewriting stages SR and PA, the coarse plan only decides whether a stage is applied, without specifying the rewrite goal or a chunk-specific instruction, so the rewriting stays generic rather than adaptive. To address both, we evolve each coarse plan by grounding it in actual tool-model execution. 

For a chunk marked for cleaning, we run the selected stages in order and verify each one with a verifier LLM, removing any stage that brings little or harmful change. We remove NP when it over-prunes non-noise lines. For a rewriting stage, the verifier checks whether the rewrite removes the visible noise and preserves fidelity, i.e., introduces no hallucinated facts and loses no information; we remove it when it repeatedly fails these checks or barely changes the chunk (Levenshtein similarity above $98\%$). For a failed rewrite, the verifier produces a corrective instruction and the tool model retries under it, up to $N=5$ times. For a rewrite that passes on the first try, the verifier synthesizes the instruction from the before-and-after chunk. For the tool models, we follow \citet{zhou2024programming} to train the NP model (like ProX-C) by fine-tuning Qwen3-0.6B-Base, keeping only the \texttt{remove\_lines(start, end)} operation as in Section~\ref{sec:framework}, and use the off-the-shelf Qwen3-4B (non-thinking) as the SR and PA model. Details are in Appendix~\ref{app:details_fine}.

\paragraph{Fine-tuning}
The procedure above yields about $300$K pairs of training data. We use them to supervised fine-tune (SFT) the orchestrator from Qwen3-1.7B-Base: given a chunk as input, the model is trained to produce its data curation pipeline, that is, whether to drop or untouch the chunk, or, if it is cleaned, which of NP, SR, and PA the chunk passes through, together with the chunk-specific instruction for each SR or PA stage involved. Training hyper-parameters are provided in Appendix~\ref{app:sft_hyperparams}. We also experiment with orchestrators of different sizes; as shown in Appendix~\ref{app:orchestrator_size}, the $1.7$B model offers the best trade-off between overall performance and cost, so we adopt it as the default.

\section{Experiments}
\label{sec:exp}

\subsection{Setup}
\label{sec:main_setup}

\paragraph{Baselines}
We select baselines that cover several types of data-processing methods. The first is a rule-based method, for which we combine the heuristic curation rules from C4~\citep{raffel2020exploring}, Gopher~\citep{rae2021scaling}, and FineWeb~\citep{penedo2024fineweb}. The remaining baselines are model-based. ProX~\citep{zhou2024programming} performs both data filtering and line-level editing. For a fair comparison, we retrain ProX models on Qwen3-0.6B-Base. We further include two rewriting baselines that use an LLM to rewrite the data. RePro~\citep{yu2025repro} is a rewriter built by fine-tuning Qwen3-4B with reinforcement learning, which faithfully recycles web data. ReWire~\citep{nguyen2025recycling} uses a prompt-based approach that asks an LLM to expand the key points of the text during rewriting. To compare these methods fairly, we use Qwen3-4B (non-thinking) as the rewriting LLM for ReWire as well, which is the same model that \method uses in its rewriting stages.

\paragraph{Data Curation and Pretraining}
We use RedPajama-V2~\citep{weber2024redpajama}, a large-scale web corpus built from Common Crawl that spans multiple quality buckets. We randomly sample 40B tokens and process them with each baseline method. Note that RePro and ReWire, in their original papers, select data with a fastText classifier and mix the rewritten data with the raw data, and we leave the comparison under this mixing setup to Section~\ref{sec:exp_mixture}. In this section, for a fair comparison with the other baselines, we do not apply such mixing and instead train directly on the rewritten data. We pretrain Transformer models from scratch at three sizes, 0.5B, 1.5B, and 7B. For all methods at 0.5B and 1.5B, we set the training budget at 20B tokens, and for 7B we train on 30B tokens. Details of the model configurations and hyper-parameters are given in Appendix~\ref{app:pretrain}.

\paragraph{Evaluation}
We evaluate on 11 widely used benchmarks: ARC-Easy and ARC-Challenge~\citep{clark2018think}, MMLU~\citep{hendrycks2020measuring}, HellaSwag~\citep{zellers2019hellaswag}, WinoGrande~\citep{sakaguchi2021winogrande}, PIQA~\citep{bisk2020piqa}, CommonsenseQA~\citep{talmor2019commonsenseqa}, SIQA~\citep{sap2019social}, RACE~\citep{lai2017race}, OpenBookQA~\citep{mihaylov2018can}, and SciQ~\citep{welbl2017crowdsourcing}. Details of evaluation are given in Appendix~\ref{app:eval}. We report accuracy on each benchmark and the average across all of them. To reduce the variance of the reported numbers, for every benchmark and for the average we report the mean over the three most recently saved checkpoints (saved about every $2$B tokens).

\subsection{Main Results}
\label{sec:main_results}

% Table~\ref{tab:main} reports downstream performance for models pretrained from scratch at 0.5B, 1.5B, and 7B parameters. \method achieves the best average score at all scales, outperforming the strongest baseline by $0.9$, $1.4$, and $1.9$ points, respectively. Its advantage grows with model size, and it ranks first on most benchmarks. Among the baselines, ProX is already competitive, showing that lightweight line-level noise removal can improve pretraining data quality. By contrast, RePro and ReWire perform poorly when models are trained only on rewritten data; RePro even falls below Raw at 1.5B and 7B. 

% However, this does not mean that full rewriting is totally ineffective. On knowledge benchmarks such as ARC and MMLU, RePro and ReWire improve performance in most cases, with ReWire showing larger gains. This suggests that rewriting can make knowledge more explicit and easier to learn. However, they often underperform on language understanding benchmarks, including HellaSwag, RACE, OpenBookQA, and PIQA. Training only on synthetic text may reduce the linguistic diversity and natural discourse patterns of web data. In addition, rewriting text into structured formats can separate reasoning steps that are naturally connected in the original prose, weakening its implicit logical flow. These effects may particularly hurt tasks requiring natural language understanding and commonsense reasoning. RePro and ReWire therefore mix rewritten and original data in their original settings; we study this setting in Section~\ref{sec:exp_mixture}.

Table~\ref{tab:main} reports downstream performance for models pretrained from scratch at 0.5B, 1.5B, and 7B parameters. \method achieves the best average score at all scales, outperforming the strongest baseline respectively, with its advantage growing with model size. Among the baselines, ProX is already competitive, showing that lightweight line-level noise removal improves data quality, whereas RePro and ReWire perform poorly when trained only on rewritten data. This does not mean full rewriting is ineffective: on knowledge benchmarks such as ARC and MMLU, RePro and ReWire improve in most cases, ReWire more so, suggesting that rewriting can make knowledge more explicit. However, they often underperform on language understanding benchmarks such as HellaSwag, RACE, OpenBookQA, and PIQA, likely because training only on synthetic text reduces the linguistic diversity and natural discourse of web data, and rewriting into structured formats can break reasoning steps that are naturally connected in the original prose. RePro and ReWire mix rewritten and original data in their original settings; we study this in Section~\ref{sec:exp_mixture}.

\begin{table}[t]
\caption{Downstream performance of models pretrained from scratch at 0.5B, 1.5B, and 7B on RedPajama-V2 data curated by different methods. The best result in each column within a panel is in \textbf{bold}, and the second best is \underline{underlined}.}
\label{tab:main}
\begin{center}
\resizebox{\textwidth}{!}{
\begin{tabular}{@{}lcccccccccccc@{}}
\toprule
\textbf{Method} & \textbf{ARC-E} & \textbf{ARC-C} & \textbf{MMLU} & \textbf{HellaS.} & \textbf{RACE} & \textbf{WinoG.} & \textbf{OBQA} & \textbf{PIQA} & \textbf{CSQA} & \textbf{SIQA} & \textbf{SciQ} & \textbf{AVG} \\
\midrule
\rowcolor{panelblue}\multicolumn{13}{c}{\textbf{0.5B}} \\
\midrule
Raw & 39.39 & 23.72 & 26.67 & 34.99 & 28.36 & 49.01 & 25.87 & 65.67 & 19.63 & 37.50 & 63.13 & 37.63 \\
Rule-based & 40.97 & 25.31 & 26.55 & \underline{36.84} & 29.25 & 52.49 & \underline{28.20} & \textbf{67.34} & 19.57 & \textbf{38.69} & 61.97 & 38.83 \\
ProX & 42.42 & 25.11 & 27.01 & \textbf{37.04} & \underline{29.31} & \underline{52.67} & 26.87 & \underline{67.03} & 19.55 & \underline{38.55} & 64.47 & \underline{39.09} \\
RePro & 39.51 & 24.54 & 27.13 & 34.89 & 27.69 & \textbf{52.80} & 25.07 & 64.67 & \underline{20.48} & 37.09 & 62.10 & 37.81 \\
ReWire & \underline{42.85} & \underline{26.68} & \underline{27.48} & 34.32 & 25.61 & 51.83 & 26.13 & 63.86 & 20.17 & 37.29 & \underline{64.60} & 38.26 \\
\method & \textbf{44.75} & \textbf{27.16} & \textbf{27.60} & 36.80 & \textbf{30.18} & 51.88 & \textbf{29.20} & 66.50 & \textbf{20.56} & 38.40 & \textbf{66.90} & \textbf{39.99} \\
\midrule
\rowcolor{panelblue}\multicolumn{13}{c}{\textbf{1.5B}} \\
\midrule
Raw & 43.13 & 25.17 & 27.41 & 40.54 & 28.87 & 50.70 & 28.53 & 68.44 & 19.08 & 38.84 & \underline{67.90} & 39.87 \\
Rule-based & 43.73 & 25.28 & 27.76 & \textbf{44.17} & 30.88 & \underline{52.38} & 30.13 & \textbf{69.73} & 19.63 & \textbf{39.61} & 64.50 & 40.71 \\
ProX & \underline{46.09} & 25.17 & 28.13 & \underline{43.46} & \underline{31.39} & 51.46 & \underline{30.53} & \underline{69.13} & 19.82 & 38.95 & 67.80 & \underline{41.08} \\
RePro & 43.01 & 26.00 & 27.43 & 38.54 & 29.35 & 51.70 & 26.27 & 67.32 & \textbf{20.47} & 37.26 & 65.87 & 39.38 \\
ReWire & 45.59 & \textbf{28.50} & \underline{28.42} & 38.16 & 29.25 & 51.99 & 27.47 & 66.14 & \underline{20.01} & 36.98 & 67.77 & 40.03 \\
\method & \textbf{49.92} & \underline{27.50} & \textbf{28.86} & 43.04 & \textbf{31.77} & \textbf{53.62} & \textbf{31.40} & 69.08 & 19.93 & \underline{39.08} & \textbf{72.63} & \textbf{42.44} \\
\midrule
\rowcolor{panelblue}\multicolumn{13}{c}{\textbf{7B}} \\
\midrule
Raw & 51.50 & 27.84 & 29.67 & 52.62 & 32.70 & 54.41 & 31.67 & 72.34 & \textbf{21.76} & 40.92 & 77.30 & 44.79 \\
Rule-based & 52.41 & 29.35 & 30.35 & \textbf{56.23} & \underline{33.72} & \underline{56.46} & \underline{34.20} & 73.34 & 19.46 & 40.74 & 73.23 & 45.41 \\
ProX & 53.72 & 29.27 & 30.56 & 55.72 & 33.62 & 54.12 & 33.13 & \underline{73.61} & 19.90 & \underline{41.66} & 77.80 & \underline{45.74} \\
RePro & 49.30 & 27.45 & 30.72 & 47.35 & 32.09 & 53.80 & 31.87 & 70.77 & 20.34 & 39.76 & 72.57 & 43.27 \\
ReWire & \underline{54.12} & \underline{32.08} & \underline{31.63} & 46.28 & 31.93 & 55.80 & 33.20 & 69.33 & \underline{21.16} & 39.90 & \underline{79.60} & 45.00 \\
\method & \textbf{58.71} & \textbf{33.96} & \textbf{32.22} & \underline{55.95} & \textbf{34.77} & \textbf{57.43} & \textbf{35.07} & \textbf{73.74} & 19.98 & \textbf{42.72} & \textbf{79.73} & \textbf{47.66} \\
\bottomrule
\end{tabular}
}
\end{center}
\end{table}

\subsection{Generalization across Datasets and Settings}

\begin{figure}[t]
\centering
\includegraphics[width=\linewidth]{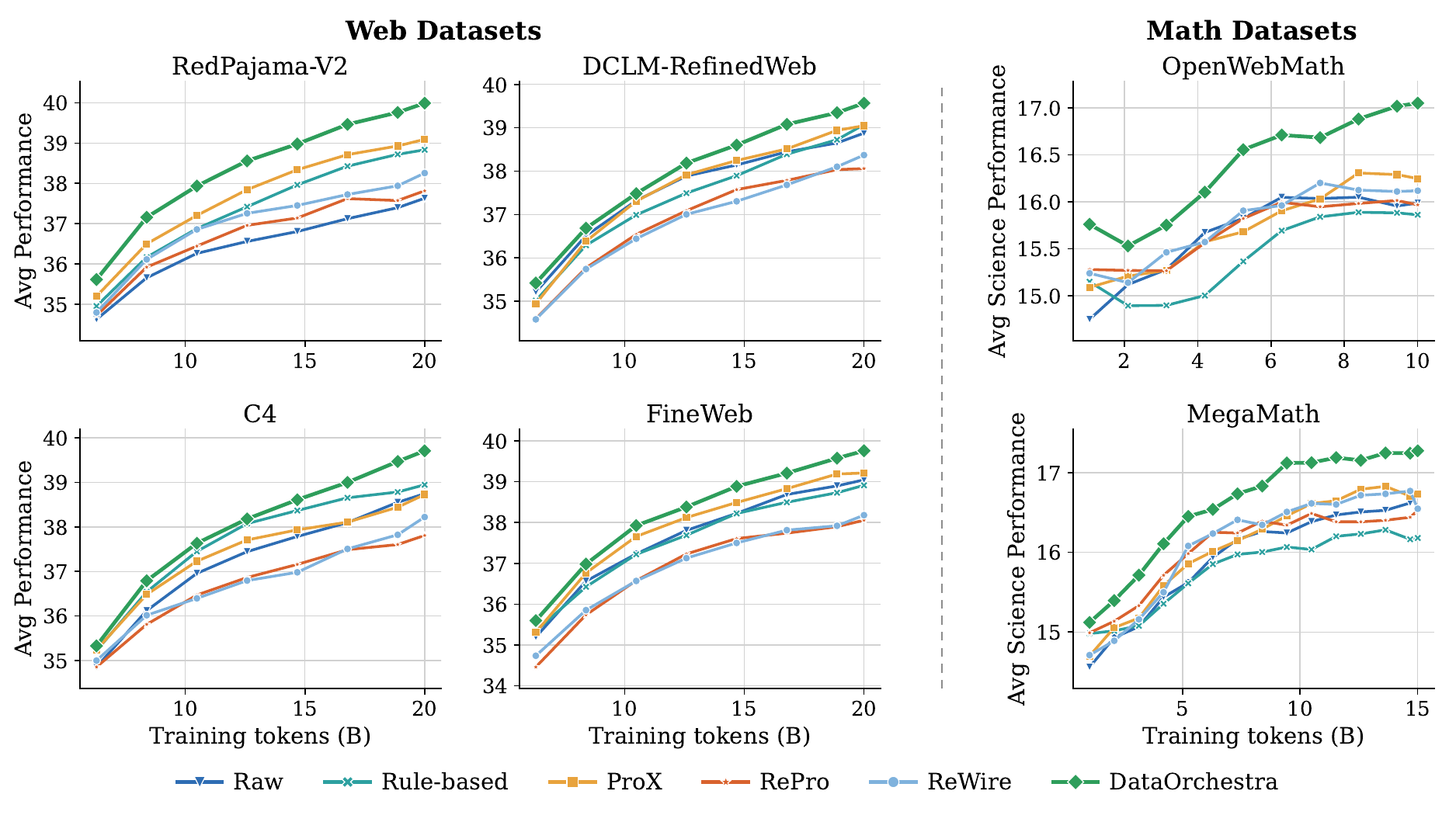}
% \caption{Average downstream performance versus training tokens for the six methods. Left: four general web datasets (0.5B models pretrained from scratch). Right: two math datasets (3B models under continued pretraining). Each curve uses the 3-checkpoint moving average of the AVG score.}
\caption{Average performance for methods including four web datasets and two math datasets.}
\label{fig:generalization}
\end{figure}

\paragraph{Additional Web Datasets}
Besides RedPajama-V2, we apply \method to three other widely used web corpora: DCLM-RefinedWeb~\citep{li2024datacomp}, C4~\citep{raffel2020exploring}, and FineWeb~\citep{penedo2024fineweb}. We follow the same setup as in Section~\ref{sec:main_setup} and pretrain models from scratch at 0.5B on the curated data. The left panel of Figure~\ref{fig:generalization} reports the results. Across all three datasets, \method gives consistent average gains over the baselines, which shows that its benefit does not depend on any particular web corpus.

\paragraph{Math Continued Pretraining}
% We also study whether \method helps on domain data, taking math as one of the most common domains. We build on Davinci-Origin-3B~\citep{qin2026data}, a fully transparent checkpoint that is pretrained on 1T tokens and has never been trained on science QA data, which makes it a clean base model for continued-pretraining experiments in the science domain. We apply all methods to two math corpora, OpenWebMath~\citep{paster2024openwebmath} and MegaMath~\citep{zhou2025megamath}. Continued pretraining on a single domain can cause catastrophic forgetting due to domain shift~\citep{luo2025empirical}, so for every method we mix the curated math data with general-domain data; the mixing details are given in Appendix~\ref{app:cpt_mixture}. We set the training budget to 10B tokens for OpenWebMath and 15B tokens for MegaMath, where both counts include the mixed-in data. We evaluate the resulting models on 9 science reasoning benchmarks, with evaluation details given in Appendix~\ref{app:eval}. As shown in the right panel of Figure~\ref{fig:generalization}, \method achieves the best average performance among all methods, confirming that it also improves data quality for domain continued pretraining. The full per-benchmark results for all datasets in this section are reported in Appendix~\ref{app:generalization_full}.
We also study whether \method helps on domain data, taking math as a representative domain. We build on Davinci-Origin-3B~\citep{qin2026data}, a fully transparent checkpoint pretrained on 1T tokens and never trained on science QA data, which makes it a clean base for continued-pretraining experiments in the science domain. We apply all methods to two math corpora, OpenWebMath~\citep{paster2024openwebmath} and MegaMath~\citep{zhou2025megamath}. Since continued pretraining on a single domain can cause catastrophic forgetting~\citep{luo2025empirical}, we mix the curated math data with general-domain data for every method (details in Appendix~\ref{app:cpt_mixture}). The training budget is 10B tokens for OpenWebMath and 15B for MegaMath, both including the mixed-in data, and we evaluate on 9 science reasoning benchmarks (Appendix~\ref{app:eval}). As shown in the right panel of Figure~\ref{fig:generalization}, \method achieves the best average performance, confirming that it also improves data quality for domain continued pretraining. Full per-benchmark results for this section are in Appendix~\ref{app:generalization_full}.

\subsection{Comparison with Mixtures of Rewritten and Raw Data}
\label{sec:exp_mixture}

\begin{figure}[t]
\centering
% --- content row: top-aligned; captions go in a separate row below ---
\begin{minipage}[t]{0.51\linewidth}
\vspace{0pt}
\centering
\includegraphics[width=\linewidth]{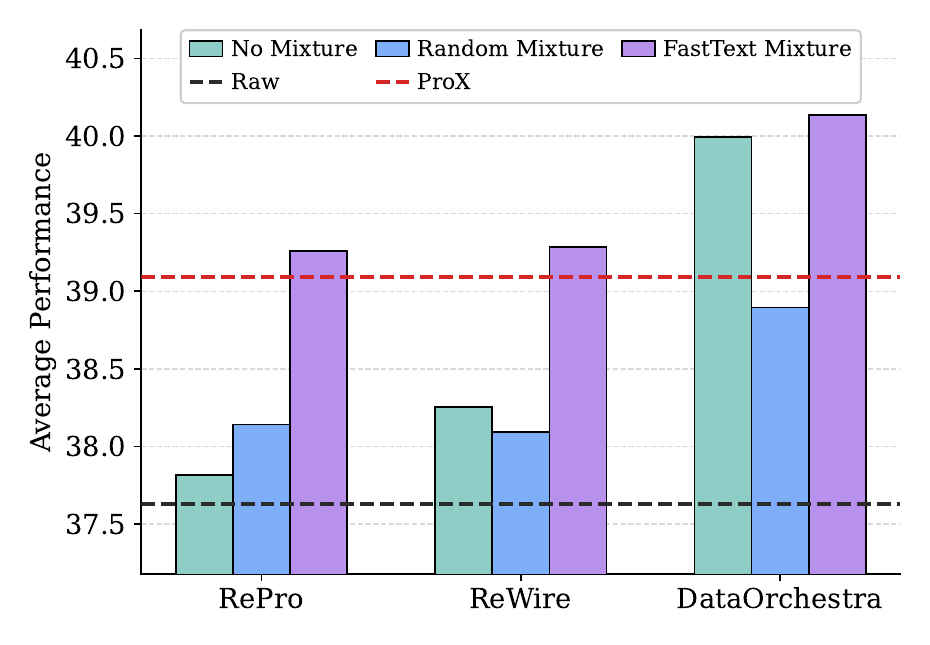}
\end{minipage}\hfill
\begin{minipage}[t]{0.47\linewidth}
\vspace{0pt}
\centering
\resizebox{\linewidth}{!}{%
\begin{tabular}{c|ccc|c|cr}
\toprule
\textbf{Drop} & \textbf{NP} & \textbf{SR} & \textbf{PA} & \textbf{Mix} & \textbf{AVG}\,$\boldsymbol{\uparrow}$ & \textbf{EFLOPs}\,$\boldsymbol{\downarrow}$ \\
\midrule
\multicolumn{4}{c|}{Raw} &            & $37.63$ & $0$ \\
\midrule
\rowcolor{grpA}\checkmark &            &            &            &            & $38.88$ & $52$ \\
\rowcolor{grpB}\checkmark & \checkmark &            &            &            & $39.09$ & $97$ \\
\rowcolor{grpC}\checkmark & \checkmark & \checkmark &            &            & $38.22$ & $572$ \\
\rowcolor{grpC}\checkmark & \checkmark & \checkmark &            & \checkmark & $39.25$ & $572$ \\
\rowcolor{grpD}\checkmark & \checkmark & \checkmark & \checkmark &            & $39.59$ & $1236$ \\
\rowcolor{grpD}\checkmark & \checkmark & \checkmark & \checkmark & \checkmark & $39.83$ & $1236$ \\
\midrule
\rowcolor{grpE}\checkmark & \multicolumn{3}{c|}{End-to-End} &            & $38.58$ & $824$ \\
\rowcolor{grpE}\checkmark & \multicolumn{3}{c|}{End-to-End} & \checkmark & $39.67$ & $824$ \\
\midrule
\rowcolor{grpF}\multicolumn{4}{c|}{\method} &            & \underline{$39.99$} & $782$ \\
\rowcolor{grpF}\multicolumn{4}{c|}{\method} & \checkmark & $\mathbf{40.13}$ & $782$ \\
\bottomrule
\end{tabular}}
\end{minipage}

\vspace{4pt}
% --- caption row: top-aligned so the two captions begin on the same line ---
\begin{minipage}[t]{0.51\linewidth}
% \caption{Effect of mixing rewritten data with raw data for the three methods that involve LLM rewriting: RePro, ReWire, and \method. For each method we compare training on the rewritten data alone (no mixture) against random and fastText mixtures with raw data. Dashed lines mark the Raw and ProX baselines.}
\caption{Effect of mixing rewritten data with raw data for methods that involve LLM rewriting.}
\label{fig:mixture}
\end{minipage}\hfill
\begin{minipage}[t]{0.47\linewidth}
\makeatletter\def\@captype{table}\makeatother
% \caption{Comparison with multi-stage data-processing pipelines. \checkmark marks the operations a fixed pipeline applies to all chunks; ``End-to-End'' replaces NP/SR/PA with a single full rewrite, and ``Mix'' indicates whether fastText-selected raw data is mixed in. EFLOPs is the compute spent on data cleaning.}
\caption{Comparison with multi-stage data-processing pipelines.}
\label{tab:multistage}
\end{minipage}
\end{figure}

In the main results (Section~\ref{sec:main_results}), RePro and ReWire perform poorly when trained only on rewritten data. Since training purely on synthetic data can cause model collapse and a loss of textual diversity, we now study mixing the rewritten data back with raw text. For each of the three rewriting methods (RePro, ReWire, and \method), we mix 10B tokens of rewritten data with 10B tokens of raw data, keeping the total budget (20B tokens) unchanged. We try two strategies: random mixing and fastText mixing which keeps the highest-scoring 10B tokens on each side using the DCLM fastText classifier~\citep{li2024datacomp}. All models are pretrained from scratch at 0.5B following Section~\ref{sec:main_setup}. Figure~\ref{fig:mixture} reports the results. Mixing helps, but depends on the quality of the raw data mixed in. FastText selection lifts both RePro and ReWire above Raw and ProX, whereas random selection is less reliable: it may mix in the low-quality part of the raw data, leaving the result near or even below Raw (e.g., ReWire). \method behaves differently: even without mixing, it already reaches the best score and surpasses every mixing configuration of the two baselines. We attribute this to the orchestrator, which naturally keeps high-quality raw chunks untouched and mixes them with the rewritten ones. Moreover, FastText mixing gives a further gain for \method. Full per-benchmark results are in Appendix~\ref{app:extra_full}.

\subsection{Comparison with Multi-stage Pipelines}
\label{sec:multi_stage}

In practice, data curation often combines several operations into a multi-stage pipeline. We therefore compare \method with such pipelines on both performance and efficiency, following Section~\ref{sec:main_setup}. Starting from Drop, we progressively add NP, SR, and PA, and also include an end-to-end variant that replaces the three stages with a single full rewrite (prompt in Appendix~\ref{app:e2e_prompt}). For every pipeline involving rewriting, we additionally train a version that mixes in fastText-selected raw data, as in Section~\ref{sec:exp_mixture}. All pipelines use the same tool models and prompts for NP, SR, and PA as \method. We also estimate the compute spent on data curation, measured by the FLOPs of LLM inference (details in Appendix~\ref{app:flops}). Results are shown in Table~\ref{tab:multistage}. Overall, performance improves steadily as more stages are added on top of fastText mixing. The end-to-end rewrite works reasonably well, yet still trails the full multi-stage pipeline. Among all methods, \method achieves the best result even without mixing, and mixing brings a further gain. It is also efficient, spending less compute than both the full NP/SR/PA pipeline and the end-to-end rewrite: although the orchestrator adds some cost, it routes each chunk to only the operations it needs, keeping the whole pipeline cost-effective.

\section{Analysis and Ablations}

\subsection{Analysis of Orchestrator Decisions}

\begin{figure}[t]
\centering
\begin{subfigure}[t]{0.5\textwidth}
\centering
\includegraphics[width=\linewidth]{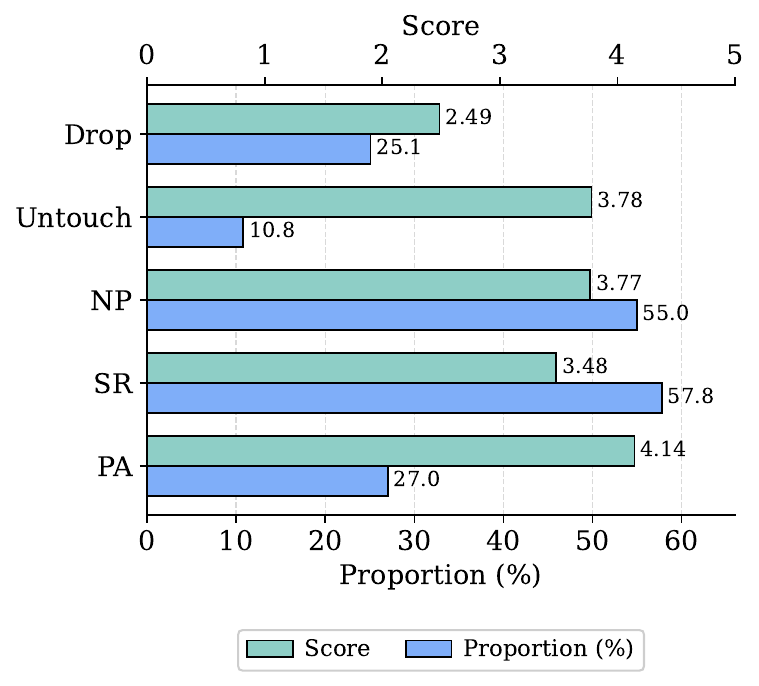}
\caption{Quality and distribution of each decision.}
\label{fig:decision_stats}
\end{subfigure}
\hfill
\begin{subfigure}[t]{0.45\textwidth}
\centering
\includegraphics[width=\linewidth]{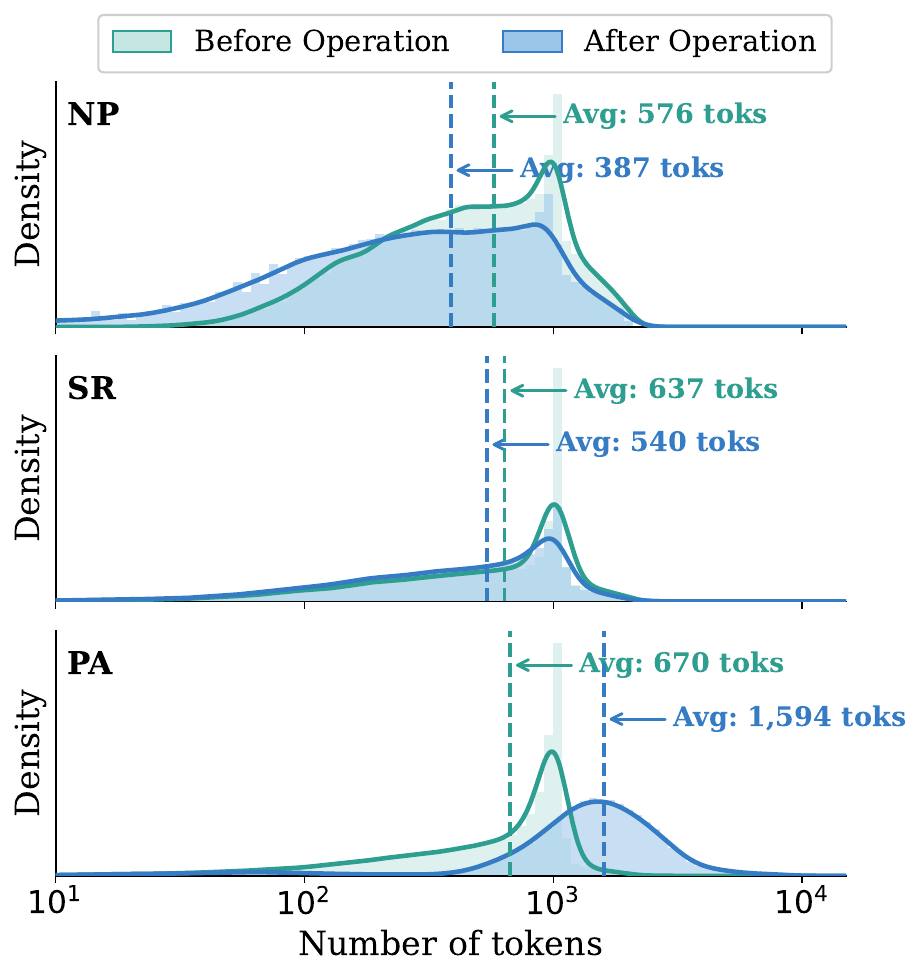}
\caption{Token distribution before/after each operation.}
\label{fig:decision_tokens}
\end{subfigure}
\caption{Analysis of the orchestrator's decisions. (a) For each operation, the average DataMan quality score of the source chunks it is applied to, and the proportion of all chunks that receive it. (b) The token-count distribution of chunks before and after NP, SR, and PA.}
\label{fig:orchestrator_decisions}
\end{figure}

\paragraph{Decisions and Data Quality}
We score every chunk with DataMan~\citep{peng2025dataman} and relate the scores to the decisions made by the orchestrator (Figure~\ref{fig:decision_stats}). Drop is applied to the lowest-scoring chunks. Among the cleaning operations, SR has the lowest source quality, since it handles chunks whose noise cannot be fixed by the lighter NP, while PA has the highest, since it targets knowledge-rich text. These trends show that the orchestrator routes each chunk in a sensible way. We also find that about 35\% of the chunks are directly dropped or left untouched in the first step, and not every cleaning chunk goes through rewriting operations, which confirms that \method saves compute by skipping unnecessary operations.

\paragraph{Token Count Changes by Operation}
Figure~\ref{fig:decision_tokens} shows the token counts before and after each operation. NP reduces tokens the most, as it removes whole lines from chunks with the most obvious surface noise. SR reduces tokens less, since it mainly repairs rather than deletes. PA instead increases tokens substantially, because it expands knowledge-rich text with additional knowledge points and reasoning steps. These changes match the design goal of each operation.

\subsection{Ablation Study of Plan Evolution and Rewriting Instruction}
\label{sec:ablation_evolution}
\begin{wrapfigure}{r}{0.5\textwidth}
\centering
\vspace{-\baselineskip}
\includegraphics[width=\linewidth]{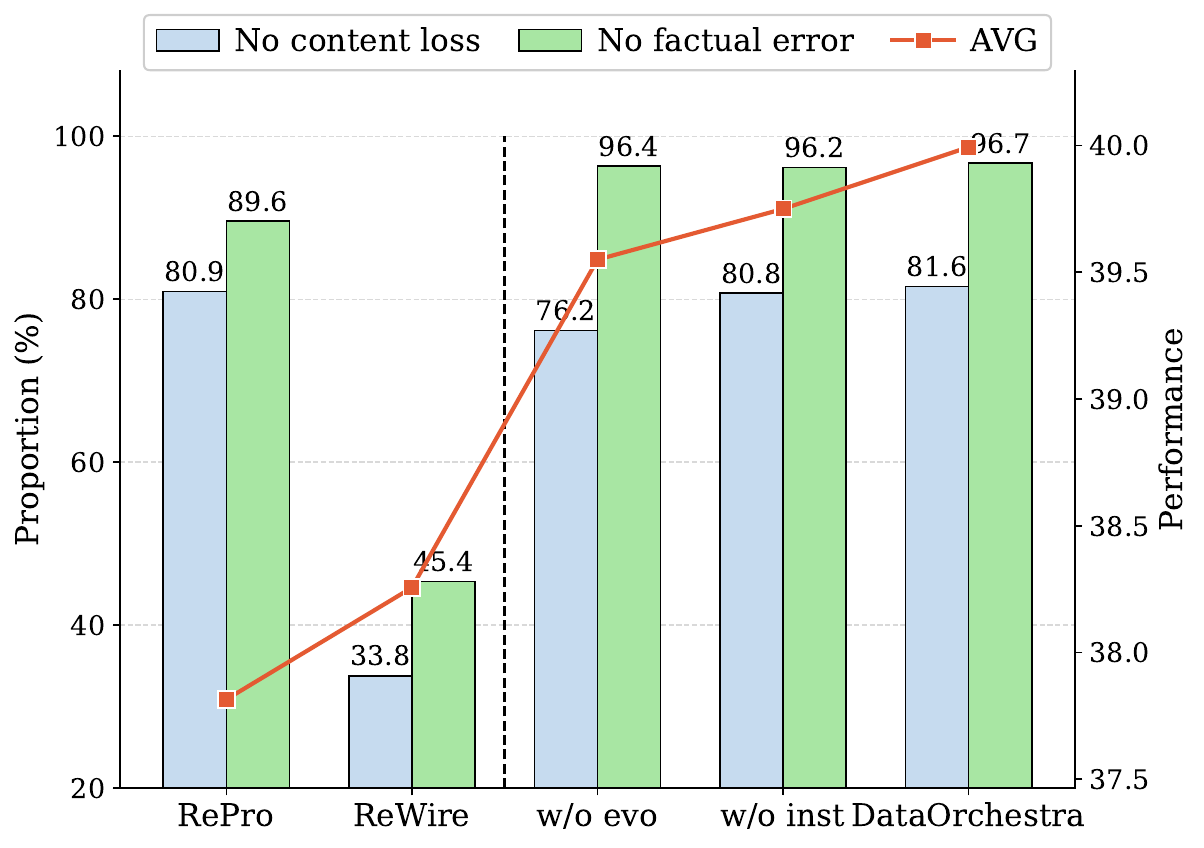}
\caption{Ablation of plan evolution and rewriting instruction.}
\label{fig:ablation_evolution}
\end{wrapfigure}

% To verify that the procedures used to train the orchestrator are effective, we conduct two ablations. \method w/o evolution removes the fine-grained plan evolution and trains the orchestrator directly on the coarse-grained plans, and therefore also generates no chunk-specific instruction. \method w/o instruction keeps the plan evolution but drops the instruction at inference time. To further analyze rewriting quality, we randomly sample 700K documents and use an LLM (Qwen3-235B-A22B-Instruct-2507) to compare each chunk before and after rewriting, judging whether any information is lost and whether any factual error is introduced. Figure~\ref{fig:ablation_evolution} reports the results, together with RePro and ReWire for comparison. Two findings emerge. First, both the plan evolution and the chunk-specific instruction improve the final performance: removing either one lowers the average score. Second, all three \method variants preserve information and avoid factual errors far better than the two baselines. RePro is relatively safe, as it is trained with reinforcement learning toward faithful rewriting, whereas ReWire, encouraged to expand the text freely, loses information and introduces incorrect knowledge more often. Interestingly, this free expansion also gives ReWire an apparent edge on some knowledge benchmarks, even though it is less faithful to the original text.

To verify that the procedures for training the orchestrator are effective, we conduct two ablations. \method w/o evolution removes the fine-grained plan evolution, training the orchestrator directly on the coarse-grained plans and thus generating no chunk-specific instruction. \method w/o instruction keeps the evolution but drops the instruction at inference time. To further assess rewriting quality, we sample 700K documents and use a teacher LLM to compare each chunk before and after rewriting, judging whether information is lost or factual errors are introduced (the evaluation prompt is given in Appendix~\ref{app:prompt_eval}). Figure~\ref{fig:ablation_evolution} reports the results, with RePro and ReWire for comparison. First, both the plan evolution and the chunk-specific instruction improve performance, as removing either lowers the average score. Second, all three \method variants preserve information and avoid factual errors far better than the baselines. RePro is relatively safe, being trained with reinforcement learning toward faithful rewriting, whereas ReWire, encouraged to expand text freely, loses information and adds incorrect knowledge more often. Interestingly, this free expansion also gives ReWire an apparent edge on some knowledge benchmarks despite being less faithful.

% \section{Limitations}

\section{Conclusion}
In this paper, we present \method, a framework that orchestrates a per-example data curation pipeline for LLM pretraining. A small orchestrator model decides, for each chunk of data, whether to drop, keep, or clean it, and further selects which cleaning stages to apply and generates a chunk-specific instruction for each rewriting step. Experiments show that \method consistently outperforms both individual data-processing methods and fixed multi-stage pipelines, while saving compute by skipping unnecessary operations.

\bibliography{ref}

\newpage

\appendix
\section*{Appendix}

\section{Details of Orchestrator Model Training}

\subsection{Details of Data Collection}
\label{app:data_collection}

Our seed pool is drawn from six widely used public pretraining corpora, covering a general domain and a science domain. The acquisition protocol for each source is detailed below.

\paragraph{General Domain}
\begin{itemize}
  \item \textbf{RedPajama-V2} A large multilingual CommonCrawl derivative hosted by Together. We restrict the selection to the English partition and apply explicit random subsampling: from the full set of English document groups---each identified by a snapshot/partition key such as \texttt{2014-15/0000}---we draw a uniform random sample of $1{,}000$ groups. For every sampled group we retrieve all three quality buckets (\texttt{head}, \texttt{middle}, \texttt{tail}) so that the resulting pool spans the full quality spectrum rather than being biased toward any single bucket.

  \item \textbf{DCLM-RefinedWeb} A CommonCrawl-derived corpus released through the DataComp effort. We retrieve a single global shard, \texttt{global-shard\_03\_of\_10}, from the public S3 prefix \nolinkurl{s3://commoncrawl/contrib/datacomp/DCLM-refinedweb/}, transferring all ten of its constituent local shards with the high-throughput \texttt{s5cmd} client. No subsampling is applied within the global shard.

  \item \textbf{C4} The Colossal Clean Crawled Corpus, in its English configuration from the HuggingFace Hub. We use the full set of obtained shards without further subsampling.

  \item \textbf{FineWeb} A deduplicated, quality-filtered CommonCrawl corpus from the HuggingFace Hub. We use the full set of obtained shards without further subsampling.
\end{itemize}

\paragraph{Science Domain}
\begin{itemize}
  \item \textbf{OpenWebMath} A mathematics-focused corpus built from CommonCrawl HTML. Its construction filters documents to retain those that are English, mathematically substantive, and of high quality, with particular care taken to faithfully extract LaTeX expressions and to suppress boilerplate relative to generic web corpora. We download the full dataset from the HuggingFace Hub.

  \item \textbf{MegaMath} A large-scale, open mathematical pretraining corpus curated from diverse math-focused sources, comprising re-extracted mathematical web documents, recalled math-related code, and synthetic data, for a total of $371$B tokens. Among its constituent subsets, \texttt{megamath-web} is the web portion, re-extracted from CommonCrawl with math-oriented HTML processing followed by fastText-based filtering and deduplication. We download the complete \texttt{megamath-web} subset from the HuggingFace Hub.
\end{itemize}

From each collected corpus we randomly sample $2.5$B tokens for Orchestrator training, with the sole exception of OpenWebMath, from which we sample $0.5$B tokens owing to its smaller total size. From the remaining documents of each corpus we then sample a further $40$B tokens for the downstream pretraining experiments in Section~\ref{sec:exp}; for OpenWebMath, whose remainder falls short of $40$B, we instead take all of its documents not used for Orchestrator training. The two splits are disjoint, so that no document seen during Orchestrator training reappears in the pretraining experiments. After the DataMan-based scoring and low-quality upsampling described in Section~\ref{sec:orchestrator_model_training}, the Orchestrator training split comprises $160$K documents in total; its per-corpus document counts are reported in Table~\ref{tab:orchestrator_doc_counts}.

\begin{table}[h]
\begin{center}
  \begin{tabular}{lr}
    \toprule
    \textbf{Corpus} & \textbf{\# Documents} \\
    \midrule
    RedPajama-V2     & $30{,}000$ \\
    DCLM-RefinedWeb  & $30{,}000$ \\
    C4               & $30{,}000$ \\
    FineWeb          & $30{,}000$ \\
    OpenWebMath      & $20{,}000$ \\
    MegaMath         & $20{,}000$ \\
    \midrule
    \textbf{Total}   & $160{,}000$ \\
    \bottomrule
  \end{tabular}
\end{center}
  \caption{Per-corpus document counts in the Orchestrator training split.}
  \label{tab:orchestrator_doc_counts}
\end{table}

\subsection{Prompts for Coarse-grained Plan Generation}
\label{app:prompt_coarse_plan}
% 2 prompts

The coarse-grained plan is produced in two steps. The first step decides whether to drop the chunk, using the quality-gate prompt shown in Figure~\ref{fig:prompt_drop}. The second step routes each kept chunk through the three cleaning stages, making a binary True/False judgment on whether it needs noise pruning, surface rectification, and pedagogical augmentation, using the prompt shown in Figure~\ref{fig:prompt_route}.

\begin{figure}[h!]
\begin{center}
\begin{tikzpicture}
    \node[
        draw=boxstroke,
        line width=1pt,
        rounded corners=10pt,
        fill=boxfill,
        inner sep=5pt,
        align=left,
        text width=0.9\linewidth
    ] (dropprompt) {
        \textbf{\footnotesize System Prompt}\par\vspace{3pt}
        \begin{Verbatim}[fontsize=\promptfont,breaklines=true,breakindent=0pt,breaksymbolleft={}]
You are a pretraining data quality gate. Your task is to assess whether a document chunk contains value as language model pretraining data, assuming it will go through a multi-stage cleaning pipeline (noise pruning -> surface rectification -> pedagogical augmentation) after your assessment.

Do NOT judge the chunk by its current surface quality -- a noisy chunk with valuable content underneath is still worth keeping. Instead, evaluate whether meaningful content can be recovered through reasonable cleaning effort.

## Important:
The downstream cleaning pipeline can perform substantial format rectification -- removing boilerplate lines, stripping conversational filler, reformatting structure, etc. Your assessment should therefore focus primarily on the intellectual and linguistic VALUE of the underlying content, not on how much surface noise currently surrounds it.

## Document Format
The chunk is wrapped in [DOC] / [/DOC] tags.
Note: chunks are split by token limits, so the beginning and end may be mid-sentence or mid-paragraph. This is a normal chunking artifact -- do not penalise truncation at chunk boundaries.

## Scoring Rubric (additive 5-point)
Points are accumulated based on the satisfaction of each criterion. Each criterion strictly requires all preceding criteria to be satisfied -- if a criterion is not met, no further points are awarded.

- Add 1 point if substantive human-authored content exists. The chunk contains text that goes beyond minimal reactive expressions, navigational chrome, or structural boilerplate. The content must convey information, construct an argument, pose a substantive question, tell a narrative, or explain a concept. Short phatic utterances, user-generated micro-posts that carry no transferable linguistic signal, purely templated metadata (download links, file format listings, site navigation, SEO slugs, cookie/ad banners), and machine-generated or auto-populated listings do NOT satisfy this criterion. Pornographic, sexually explicit, fetishistic, or gratuitously violent content also does NOT satisfy this criterion regardless of length or coherence.
- Add another point if the substantive content carries topical coherence around a subject that constitutes learnable knowledge or meaningful linguistic structure. The reader should be able to identify a discernible theme -- a topic being discussed, a question being investigated, a story being told -- not merely a collection of fragments that share a surface category. Coherence around purely structural or navigational purposes does not count.
- Award a third point if the content, beneath any surface noise, is intellectually intact -- the ideas, reasoning, or narrative are not fundamentally broken. Fail this only if the core content itself is garbled, machine-generated gibberish masquerading as real text, or corrupted beyond semantic recovery. Surface-level noise of any kind does NOT cause failure here.
- Grant a fourth point if the recoverable content carries genuine training value -- it provides factual information, coherent reasoning, narrative depth, technical explanation, well-posed questions with meaningful setup, or other linguistically rich material. Generic filler, SEO repetition, trivially templated text, and shallow listicles without original synthesis do NOT qualify. A well-articulated question carries value even without an answer.
- Bestow a fifth point if the content exhibits high information density or domain value -- specialised knowledge, well-structured argumentation, educational depth, or rare high-quality material that would clearly enrich corpus diversity or quality.
        \end{Verbatim}
        \vspace{1pt}\hdashrule{\linewidth}{0.5pt}{2pt}\par\vspace{1pt}
        \textbf{\footnotesize User Prompt}\par\vspace{2pt}
        \begin{Verbatim}[fontsize=\promptfont,breaklines=true,breakindent=0pt,breaksymbolleft={},commandchars=\@\!\?]
[DOC]
@hlchunk!{chunk}?
[/DOC]

## Output Format
Respond in the following format without any other part.
```
## Analysis
<Briefly justify your total score. Focus on what content is recoverable and what makes it worth or not worth cleaning>

## Result
<total points, an integer from 0 to 5>
```
        \end{Verbatim}
    };
\end{tikzpicture}
\end{center}
\caption{Prompt for the drop-decision quality gate in coarse-grained plan generation.}
\label{fig:prompt_drop}
\end{figure}

\begin{figure}[h!]
\begin{center}
\begin{tikzpicture}
    \node[
        draw=boxstroke,
        line width=1pt,
        rounded corners=10pt,
        fill=boxfill,
        inner sep=5pt,
        align=left,
        text width=0.9\linewidth
    ] (routeprompt) {
        \textbf{\footnotesize System Prompt}\par\vspace{3pt}
        \begin{Verbatim}[fontsize=\promptfont,breaklines=true,breakindent=0pt,breaksymbolleft={}]
You are a pretraining-data cleaning router. The document chunk you see has ALREADY passed an upstream drop gate -- it is worth keeping as pretraining data. Your job is to answer THREE yes/no sub-questions about which downstream tool models, if any, should run on this chunk.

The chunk is wrapped in [DOC] / [/DOC] tags. Chunks are split by token limits, so the beginning and end may be mid-sentence or mid-paragraph -- this is a normal chunking artifact, not a defect.

## The three sub-questions

### (1) noise_pruning
**Does this chunk contain any ENTIRE LINES of structural / boilerplate noise that can be removed cleanly by whole-line deletion?**
`true` when one or more whole lines fall into any of these categories:
- site UI / navigation / breadcrumbs / menu / sidebar / footer lines
- social chrome, share buttons, comment-form scaffolding
- reference / bibliography / citation blocks; book-card metadata (ISBN, page counts, standalone "Title: ..." / "Author: ..." / publication dates)
- copyright / legal / cookie / disclaimer boilerplate; acknowledgments
- page headers / footers, RSS-style page titles, feed markers
- tracking tokens, ad chrome, download/login links separated as their own lines
- other format-level noise that can be cleanly removed by whole-line deletion ...
`false` when:
- noise (if any) is inline INSIDE otherwise substantive lines -- line-level deletion would be too coarse
- the chunk is already clean of whole-line noise

### (2) surface_rectification
**Does the text carry disorganized FORMAT issues that requires REWRITING to fix?**
`true` when one or more of these applies:
- broken text flow from OCR / PDF / HTML extraction (mid-word hyphenation, sentences split across many short line)
- structured content that has been shattered into short fragmented lines -- a table whose rows / columns are scattered across many isolated cells, an equation whose terms are split line by line, a code block whose statements are broken across single-character fragments, etc
- mojibake / character-encoding artifacts / raw HTML entities mixed into prose
- whitespace or indentation that has lost the original block / paragraph structure
- any other format-level irregularities that need rephrasing to improve quality ...
`false` when the remaining text reads cleanly as-is.

### (3) pedagogical_augmentation
**Is this chunk knowledge-dense or reasoning-intensive?**
This is a TEXT-TYPE classification.
`true` for text types which contain rich knowledge or deep reasoning like: scientific / research papers, technical documentation, mathematical or algorithmic derivations, in-depth tutorials with non-trivial reasoning, textbook material, domain-expert analyses, code with non-trivial logic, etc.
`false` for general-purpose text types like: news articles, casual blogs, narrative prose, forum discussions, product / business / restaurant listings, encyclopedic summaries, FAQ / customer-service text, recipes, schedules, lyrics, the long tail of general-purpose web text. Even if the writing is good, if the genre is general-purpose, the answer is `false`.

## Output format
Respond with EXACTLY one JSON object and nothing else -- no preamble, no trailing commentary, no markdown headers outside the JSON. If you must use a code fence, use ```json ... ```.
```json
{
  "noise_pruning":            {"reasoning": "<1-2 sentences>", "value": true | false},
  "surface_rectification":    {"reasoning": "<1-2 sentences>", "value": true | false},
  "pedagogical_augmentation": {"reasoning": "<1-2 sentences>", "value": true | false}
}
```
Each `value` is exactly the boolean `true` or `false`. Each `reasoning` is 1-2 short sentences grounded in concrete observations about the chunk.
        \end{Verbatim}
        \vspace{1pt}\hdashrule{\linewidth}{0.5pt}{2pt}\par\vspace{1pt}
        \textbf{\footnotesize User Prompt}\par\vspace{2pt}
        \begin{Verbatim}[fontsize=\promptfont,breaklines=true,breakindent=0pt,breaksymbolleft={},commandchars=\@\!\?]
## Input
[DOC]
@hlchunk!{chunk}?
[/DOC]
        \end{Verbatim}
    };
\end{tikzpicture}
\end{center}
\caption{Prompt for the cleaning-stage routing in coarse-grained plan generation.}
\label{fig:prompt_route}
\end{figure}

\subsection{Details of Fine-grained Plan Evolution}
\label{app:details_fine}

\subsubsection{Prompts for Tool LLMs}
\label{app:prompt_tool}

Each cleaning stage is executed by its own tool LLM. For noise pruning (NP), we follow ProX~\citep{zhou2024programming} and fine-tune a small model dedicated to this stage. Since the behavior is learned during fine-tuning, NP does not need an elaborate prompt: as shown in Figure~\ref{fig:prompt_np}. For surface rectification (SR), our prompt is a lightly modified version of RePro~\citep{yu2025repro}, shown in Figure~\ref{fig:prompt_sr}. For pedagogical augmentation (PA), the prompt follows the same template as SR but targets pedagogical enrichment rather than surface repair, and is shown in Figure~\ref{fig:prompt_pa}.

\begin{figure}[h!]
\begin{center}
\begin{tikzpicture}
    \node[
        draw=boxstroke,
        line width=1pt,
        rounded corners=10pt,
        fill=boxfill,
        inner sep=5pt,
        align=left,
        text width=0.9\linewidth
    ] (npprompt) {
        \textbf{\footnotesize System Prompt}\par\vspace{3pt}
        \begin{Verbatim}[fontsize=\promptfont,breaklines=true,breakindent=0pt,breaksymbolleft={}]
You are an excellent noise pruning model for pretraining data cleaning.
        \end{Verbatim}
        \vspace{1pt}\hdashrule{\linewidth}{0.5pt}{2pt}\par\vspace{1pt}
        \textbf{\footnotesize User Prompt}\par\vspace{2pt}
        \begin{Verbatim}[fontsize=\promptfont,breaklines=true,breakindent=0pt,breaksymbolleft={},commandchars=\@\!\?]
[DOC]
@hlchunk!{chunk}?
[/DOC]
        \end{Verbatim}
    };
\end{tikzpicture}
\end{center}
\caption{Prompt for the noise pruning (NP) tool LLM.}
\label{fig:prompt_np}
\end{figure}

\begin{figure}[h!]
\begin{center}
\begin{tikzpicture}
    \node[
        draw=boxstroke,
        line width=1pt,
        rounded corners=10pt,
        fill=boxfill,
        inner sep=5pt,
        align=left,
        text width=0.9\linewidth
    ] (srprompt) {
        \textbf{\footnotesize System Prompt}\par\vspace{3pt}
        \begin{Verbatim}[fontsize=\promptfont,breaklines=true,breakindent=0pt,breaksymbolleft={},commandchars=\@\!\?]
You are a **surface rectification** specialist for pretraining data cleaning. Your job is to fix surface-level issues -- formatting damage, structural breakage, and disorganized text -- through careful rewriting. You must NOT alter the original meaning, introduce new information, or add any knowledge that is not already present in the text. This is a meaning-preserving rewrite, not a knowledge-adding one.

## Document Format
Each given document chunk is wrapped in [DOC] / [/DOC] tags.
Note: chunks are split by token limits, so the beginning and end may be mid-sentence or mid-paragraph. This is a normal chunking artifact -- do not attempt to complete or extend truncated text at chunk boundaries.

## Instructions
Read the chunk thoroughly, then paraphrase it in high-quality and clear English following these rules:
- Delete clearly irrelevant content:
  - Website headers, navigation bars, or menu items (e.g., "Home | About | Contact")
  - Unrelated HTTP links (e.g., ads, trackers, developer tools)
  - Generic footers (e.g., contact info, privacy policies, unsubscribe links)
  - Empty lines or decorative elements (e.g., "---")
- Repair damaged structured content:
  - Repair damaged tables, diagrams, formulas, or code blocks that appear as consecutive lines of isolated words, single characters, or short fragments. Use surrounding context to reconstruct the intended structure.
  - Limited inference from surrounding context is permitted in this case.
- Preserve all content that is relevant and meaningful:
  - Informative or independently useful
  - Related to the topic, even tangentially
  - Provides context, background, or supporting value
  - Includes technical terms, key concepts, factual details, reasoning, and examples
- Handle mixed-relevance sentences carefully:
  - Remove only the irrelevant fragment if the rest remains coherent
  - Delete the whole sentence if the remainder loses meaning
- Do not alter meaningful content unnecessarily:
  - Only delete or modify when content is clearly meaningless or off-topic
  - Preserve the original structure, logic, and depth of the text
- Do not add explanations, notes, assumptions, or claims not found in the original text

You may optionally consult the following expert rewriting suggestion.
[HINT]
@hlchunk!{instruction}?
[/HINT]
        \end{Verbatim}
        \vspace{1pt}\hdashrule{\linewidth}{0.5pt}{2pt}\par\vspace{1pt}
        \textbf{\footnotesize User Prompt}\par\vspace{2pt}
        \begin{Verbatim}[fontsize=\promptfont,breaklines=true,breakindent=0pt,breaksymbolleft={},commandchars=\@\!\?]
## Input
[DOC]
@hlchunk!{chunk}?
[/DOC]

## Output Format
output ONLY the refined chunk as plain text (no introductory or concluding remarks, no [DOC]/[/DOC] or [HINT]/[/HINT] wrappers).
        \end{Verbatim}
    };
\end{tikzpicture}
\end{center}
\caption{Prompt for the surface rectification (SR) tool LLM.}
\label{fig:prompt_sr}
\end{figure}

\begin{figure}[h!]
\begin{center}
\begin{tikzpicture}
    \node[
        draw=boxstroke,
        line width=1pt,
        rounded corners=10pt,
        fill=boxfill,
        inner sep=5pt,
        align=left,
        text width=0.9\linewidth
    ] (paprompt) {
        \textbf{\footnotesize System Prompt}\par\vspace{3pt}
        \begin{Verbatim}[fontsize=\promptfont,breaklines=true,breakindent=0pt,breaksymbolleft={},commandchars=\@\+\=]
You are a **pedagogical augmentation** specialist for pretraining data cleaning. Your job is to enhance the pedagogical quality of document chunks by adding explanatory depth, clarifying reasoning, and lowering cognitive load for readers.

## Document Format
Each given document chunk is wrapped in [DOC] / [/DOC] tags.
Note: chunks are split by token limits, so the beginning and end may be mid-sentence or mid-paragraph. This is a normal chunking artifact -- do not attempt to complete or extend truncated text at chunk boundaries.

## What to Augment
Augment the text by making implicit reasoning explicit (the goal, the strategy, and why it fits), demystifying jargon and notation with intuition-building analogies and examples before the formal explanation, illustrating abstract ideas with concrete cases, drawing contextual bridges to the broader field and foundational concepts, and anticipating learner confusion by addressing likely questions and flagging common misconceptions -- all woven naturally into the narrative like a mentor's margin notes.
ALL data, formulas, definitions, theories, experimental results, and logical arguments must be preserved without altering their meaning. Your additions must clarify, never contradict. Do not directly generate an explanation based on the original text; it must be a self-consistent text that retains elements from the original text!

You may optionally consult the following expert rewriting suggestion.
[HINT]
@hlchunk+{instruction}=
[/HINT]
        \end{Verbatim}
        \vspace{1pt}\hdashrule{\linewidth}{0.5pt}{2pt}\par\vspace{1pt}
        \textbf{\footnotesize User Prompt}\par\vspace{2pt}
        \begin{Verbatim}[fontsize=\promptfont,breaklines=true,breakindent=0pt,breaksymbolleft={},commandchars=\@\!\?]
## Input
[DOC]
@hlchunk!{chunk}?
[/DOC]

## Output Format
output ONLY the fully augmented chunk as a self-contained educational text (plain text only, no introductory or concluding remarks, no [DOC]/[/DOC] or [HINT]/[/HINT] wrappers).
        \end{Verbatim}
    };
\end{tikzpicture}
\end{center}
\caption{Prompt for the pedagogical augmentation (PA) tool LLM.}
\label{fig:prompt_pa}
\end{figure}

\subsubsection{Prompts for Verifier LLMs}
\label{app:prompt_verifier}

During fine-grained plan evolution, every executed stage is checked by a verifier LLM before it is committed to the plan, and we use two verifier prompts. The NP verifier, shown in Figure~\ref{fig:prompt_npverify}, receives the chunk before and after pruning, both line-numbered under the same scheme, and returns a \texttt{keep}/\texttt{revert} verdict; it reverts the NP step only when the deleted lines carry substantive content. The rewrite verifier, shown in Figure~\ref{fig:prompt_rewriteverify}, is shared by SR and PA: it compares the before and after versions, scores the rewrite on a three-item rubric (no content loss, no factual error, no surface noise), and writes a chunk-specific instruction that drives the next rewriting attempt. To validate the reliability of these verifiers, we conduct a manual review of a sample of their verdicts and find over $83\%$ agreement with human judgments.

\begin{figure}[h!]
\begin{center}
\begin{tikzpicture}
    \node[
        draw=boxstroke,
        line width=1pt,
        rounded corners=10pt,
        fill=boxfill,
        inner sep=5pt,
        align=left,
        text width=0.9\linewidth
    ] (npverify) {
        \textbf{\footnotesize System Prompt}\par\vspace{3pt}
        \begin{Verbatim}[fontsize=\promptfont,breaklines=true,breakindent=0pt,breaksymbolleft={}]
You are an **NP verifier** for pretraining-data cleaning. A noise-pruning specialist has deleted some lines from a document chunk. Your job is to decide whether those deletions removed any substantive content -- if so, the deletions must be reverted; otherwise they are kept.

## Input format
You will receive two views of the same chunk, both line-numbered with the SAME numbering scheme so deletions are easy to locate:
- **BEFORE NP** -- the chunk before pruning. Every line carries a [NNN] prefix.
- **AFTER NP** -- the chunk after pruning. Lines that were removed have been replaced by a literal [NNN] <removed> marker so the line-number alignment with BEFORE is preserved.

Both chunks are wrapped in [DOC] / [/DOC]. Chunks are token-bounded slices; mid-sentence ends at the chunk boundary are normal -- do not treat them as defects.

## What counts as substantive content
A removed line is **substantive** if it carries factual content, definitions, named entities, code, equations, table data, narrative body, claims, or any other information a reader would actually learn from. Substantive content survives even when it sits next to surface noise.

A removed line is **NOT substantive (i.e. legitimate noise)** if it is:
- site UI / navigation / breadcrumbs / menus / footers / share buttons;
- bibliographic or catalog metadata (ISBN, page counts, publication dates, standalone author/title lines, journal/citation chrome);
- review/rating tuples without actual review text (e.g. "Amy B. - 2009 - 5 stars");
- page titles, RSS-style headers, feed markers, copyright/legal boilerplate;
- FAQ / marketing / sign-offs ("P.S.", "<3 Stacy", "Stay tuned...");
- empty lines, decorative dividers ("---"), tracking tokens, ad chrome.

## Verdict rule
- `keep`   -- removed lines are legitimate noise (or trivially absent). The NP deletions stand.
- `revert` -- removed lines carry substantive content. The whole NP step is reverted.

Be calibrated, not preservationist. Before issuing `revert`, ask yourself honestly: do the deleted lines actually carry educational value for a language model -- would training on those lines genuinely make the model better at understanding language, knowledge, or reasoning? Most things noise-pruning typically removes (site chrome, catalog metadata, listings without prose, sign-offs, page titles) carry near-zero pretraining value even though they ARE original text from the chunk -- deleting them is the whole point of cleaning, not a defect. A small amount of "real" content being removed alongside obvious noise is NOT enough to revert; only choose `revert` when the deletions strip away genuinely important substantive content. When in doubt between "this removal is fine" and "this removal lost a tiny bit", lean `keep`.

## Output format
Respond with EXACTLY one JSON object and nothing else. If you must use a code fence, use ```json ... ```.
```json
{
  "reasoning": "<1-2 sentences>",
  "verdict": "keep" | "revert"
}
```
        \end{Verbatim}
        \vspace{1pt}\hdashrule{\linewidth}{0.5pt}{2pt}\par\vspace{1pt}
        \textbf{\footnotesize User Prompt}\par\vspace{2pt}
        \begin{Verbatim}[fontsize=\promptfont,breaklines=true,breakindent=0pt,breaksymbolleft={},commandchars=\@\!\?]
## BEFORE NP
[DOC]
@hlchunk!{before_chunk}?
[/DOC]

## AFTER NP
[DOC]
@hlchunk!{after_chunk}?
[/DOC]
        \end{Verbatim}
    };
\end{tikzpicture}
\end{center}
\caption{Prompt for the noise pruning (NP) verifier LLM.}
\label{fig:prompt_npverify}
\end{figure}

\begin{figure}[p]
\begin{center}
\begin{tikzpicture}
    \node[
        draw=boxstroke,
        line width=1pt,
        rounded corners=10pt,
        fill=boxfill,
        inner sep=5pt,
        align=left,
        text width=0.9\linewidth
    ] (rewriteverify) {
        \textbf{\footnotesize System Prompt}\par\vspace{3pt}
        \begin{Verbatim}[fontsize=\promptfontsmall,breaklines=true,breakindent=0pt,breaksymbolleft={}]
You are a **rewrite verifier and instruction author** for pretraining-data cleaning. A cleaning specialist has produced an AFTER version of a document chunk from its BEFORE version -- through format-fixing rewriting, or pedagogical augmentation that unpacks implicit reasoning / demystifies jargon / lowers cognitive load, or both. The specialist must not alter the meaning of the original material. You receive two versions of the SAME chunk and must (1) score the rewrite on a 3-item rubric AND (2) write a chunk-specific instruction that drives the next pass.

## Input format
- **BEFORE** -- the chunk before the cleaning pass.
- **AFTER**  -- the chunk after the cleaning pass.

Both BEFORE and AFTER are wrapped in [DOC] / [/DOC]. Chunks are token-bounded slices; mid-sentence ends at chunk boundaries are normal.

## Scoring rubric
For each item write `reasoning` first (1-2 sentences), then `score` (the integer `0` or `1` -- `1` means "rubric item passes / no problem").

`no_content_loss` -- Every substantive piece of content in the BEFORE chunk survives in the AFTER chunk. "Substantive" covers facts, numerical values, named entities, definitions, examples, claims, reasoning steps, code, data, etc. Reformatting and interleaving entailed explanatory additions are fine. Removing genuine noisy information is fine and expected -- site navigation, ads, social-share UI, cookie banners, repeated headers/footers, copyright/legal boilerplate, etc. The failure modes to catch: (1) originally-present important information has been deleted, or replaced by a higher-level gloss/summary rather than preserved (e.g. exact numbers, formulas, terminology, quoted phrasing, or code dropped in favour of a paraphrase); (2) the AFTER chunk has been turned into commentary ABOUT the original -- an explanation, description, or meta-discussion of what the BEFORE text says -- rather than remaining a faithful presentation of the text itself, so that the original's own integrity as primary content is lost even if its points are mentioned.

`no_factual_error` -- The rewrite introduced no factually incorrect or hallucinated information. Reconstructing damaged content from context, and inserting reasoning steps, intuitions, analogies, or jargon clarifications, are allowed -- but every such addition must be factually correct and entailed by the BEFORE text (or by uncontroversial domain knowledge naturally implied by it).

`no_surface_noise` -- AFTER does not still carry obvious surface-level format noise (leftover headers/footers, timestamps, social chrome, cookie banners, breadcrumbs, tracking tokens, raw HTML scaffolding, OCR line breaks the specialist did not stitch up, mojibake, scattered single-character fragments from a damaged table/equation/code block, etc.). Also penalise meta-commentary artifacts introduced by the specialist itself -- preambles or postambles like "Here is the rephrased version:", "Below is the cleaned/enriched text:" -- if any appear, score 0.

## Instruction
Always emit an `instruction` field:
- **If ANY rubric item scored 0** -- the instruction tells the next attempt how to fix the specific problems you flagged. For example, emphasizing what needs to be fixed, what to keep, what to augment, or what not to expand (if you find that expanding a part easily causes hallucinations).
- **If ALL three rubric items scored 1** -- write a detailed instruction that, read forward-looking from only BEFORE, would steer a future specialist to produce a result of the same quality.

PS:
- Your instruction MUST be grounded ONLY in what is visibly present in the BEFORE chunk. You may ONLY point at WHICH part needs WHICH kind of treatment -- you must NEVER supply the corrected text, the rewritten sentence, the reconstructed table contents, the unpacked reasoning, the definition, the analogy, the example, or the intended wording yourself. For example, "rejoin the OCR-split lines in the opening paragraph" is allowed; "the opening paragraph should read 'The system processes requests in three stages...'" is forbidden because it authors the fix / explanation instead of pointing at it.
- Be concretely chunk-specific -- never generic guidance that could fit any chunk. Bad: "fix the formatting". Good: "rejoin the OCR-split lines in the beginning and rebuild the scattered three-column table near the end into aligned rows".

Either way, the instruction must be a concrete, chunk-specific directive (usually 3-5 detailed points, at most 10 points, covering only the important issues -- do not mention trivial nits such as a single misspelling, a stray hyphen, or a missing space/comma, etc) -- never generic guidance that could fit any chunk.

## Output format
Respond with EXACTLY one JSON object and nothing else. If you must use a code fence, use ```json ... ```.
```json
{
  "no_content_loss":    {"reasoning": "<sentences>", "score": 0 | 1},
  "no_factual_error":   {"reasoning": "<sentences>", "score": 0 | 1},
  "no_surface_noise":   {"reasoning": "<sentences>", "score": 0 | 1},
  "instruction": "<clear chunk-specific instruction>"
}
```
Each `score` is exactly the integer `0` or `1`. `instruction` must be a non-empty string. Do NOT add any extra keys.
        \end{Verbatim}
        \vspace{1pt}\hdashrule{\linewidth}{0.5pt}{2pt}\par\vspace{1pt}
        \textbf{\footnotesize User Prompt}\par\vspace{2pt}
        \begin{Verbatim}[fontsize=\promptfontsmall,breaklines=true,breakindent=0pt,breaksymbolleft={},commandchars=\@\!\?]
## BEFORE
[DOC]
@hlchunk!{before_chunk}?
[/DOC]

## AFTER
[DOC]
@hlchunk!{after_chunk}?
[/DOC]
        \end{Verbatim}
    };
\end{tikzpicture}
\end{center}
\caption{Prompt for the rewrite verifier LLM, shared by the SR and PA stages.}
\label{fig:prompt_rewriteverify}
\end{figure}

\subsubsection{Fine-grained Plan Evolution Algorithm}

Algorithm~\ref{alg:evolve} summarizes the fine-grained plan evolution described in Section~\ref{sec:orchestrator_model_training} for a single chunk. We write $g_s$ for the general instruction shared by all chunks routed to a rewriting stage $s$, and $p_s$ for the chunk-specific instruction that the procedure finally attaches to a retained stage; $\mathrm{sim}(\cdot,\cdot)$ denotes the Levenshtein similarity between two chunks. Starting from the running chunk $x = c$, the procedure processes the selected stages in pipeline order: it reverts NP when it over-prunes non-noise lines, and retries each rewriting stage under verifier-provided corrective instructions, dropping a stage that repeatedly fails or barely changes the chunk and otherwise annotating it with $p_s$.

\begin{algorithm}[t]
\caption{Fine-grained Plan Evolution for one chunk}
\label{alg:evolve}
\small
\begin{algorithmic}[1]
\Require chunk $c$; selected cleaning stages $\mathcal{S} \subseteq \{\mathrm{NP}, \mathrm{SR}, \mathrm{PA}\}$ in pipeline order; verifier $V$; tool models; max retries $N$; similarity threshold $\tau$

\State $x \gets c$
\For{each stage $s \in \mathcal{S}$ in pipeline order}
  \If{$s = \mathrm{NP}$}
    \State $x' \gets \mathrm{np}(x)$
    \If{$V$ finds non-noise lines over-pruned in $x'$}
      \State drop $\mathrm{NP}$ from the plan \Comment{discard $x'$, keep $x$}
    \Else
      \State $x \gets x'$
    \EndIf
  \Else \Comment{$s \in \{\mathrm{SR}, \mathrm{PA}\}$: a rewriting stage}
    \State $\mathit{instruction} \gets g_s$; \quad $\mathit{ok} \gets \textsc{False}$
    \For{$t \gets 1$ \textbf{to} $N$}
      \State $x' \gets s(x, \mathit{instruction})$ \Comment{rewrite under the current instruction}
      \If{$V$ confirms noise removed \textbf{and} fidelity preserved}
        \State $\mathit{ok} \gets \textsc{True}$; \quad \textbf{break}
      \EndIf
      \State $p_s \gets$ corrective instruction from $V$; \quad $\mathit{instruction} \gets g_s \,\Vert\, p_s$
    \EndFor
    \If{$\neg\,\mathit{ok}$ \textbf{or} $\mathrm{sim}(x, x') > \tau$}
      \State drop $s$ from the plan \Comment{little or harmful change; discard $x'$, keep $x$}
    \Else
      \If{$t = 1$} \Comment{passed on the first attempt}
        \State $p_s \gets$ instruction synthesized by $V$ from $(x, x')$
      \EndIf
      \State annotate $s$ with $p_s$; \quad $x \gets x'$
    \EndIf
  \EndIf
\EndFor
\end{algorithmic}
\end{algorithm}

\subsection{SFT Hyper-parameters}
\label{app:sft_hyperparams}

We supervised fine-tune the Orchestrator from Qwen3-1.7B-Base on the roughly $300$K (input, plan) pairs described in Section~\ref{sec:orchestrator_model_training}. Table~\ref{tab:sft_hyperparams} lists the hyper-parameters. We train for $3$ epochs with a peak learning rate of $3{\times}10^{-5}$ under a cosine schedule that decays to $10\%$ of the peak, after a warmup over the first $3\%$ of steps. We use a global batch size of $256$ and a context length of $4096$ tokens.

\begin{table}[h]
\begin{center}
  \begin{tabular}{lr}
    \toprule
    \textbf{Hyper-parameter} & \textbf{Value} \\
    \midrule
    Base model                  & Qwen3-1.7B-Base \\
    Global batch size           & $256$ \\
    Learning rate               & $3\times10^{-5}$ \\
    Epochs                      & $3$ \\
    LR scheduler                & cosine with min LR \\
    Min LR ratio                & $0.1$ \\
    Warmup ratio                & $0.03$ \\
    Context length              & $4096$ \\
    \bottomrule
  \end{tabular}
\end{center}
  \caption{Hyper-parameters for supervised fine-tuning of the Orchestrator.}
  \label{tab:sft_hyperparams}
\end{table}

\subsection{Effect of Orchestrator Size}
\label{app:orchestrator_size}

We study how the size of the Orchestrator affects the quality of the curated data. We fine-tune three Orchestrators from Qwen3-0.6B-Base, Qwen3-1.7B-Base, and Qwen3-4B-Base on the same data, use each to curate the corpus, and pretrain a 0.5B model from scratch under the setup of Section~\ref{sec:main_setup}. Table~\ref{tab:orchestrator_size} reports the per-benchmark results. Performance improves clearly from the 0.6B to the 1.7B Orchestrator. However, scaling further to 4B does not help: its average performance is slightly below even the 0.6B model. Based on case study together with human review, we conjecture that the 4B Orchestrator tends to be more rigorous and conservative, and when generating rewriting instructions it often over-emphasizes preserving all of the original structure and information, which leads the downstream rewriting model to make more trivial edits. We therefore use the 1.7B Orchestrator by default, as it gives the best overall performance at a moderate cost.

\begin{table}[h]
\begin{center}
\resizebox{\textwidth}{!}{
\begin{tabular}{@{}lcccccccccccc@{}}
\toprule
\textbf{Orchestrator} & \textbf{ARC-E} & \textbf{ARC-C} & \textbf{MMLU} & \textbf{HellaS.} & \textbf{RACE} & \textbf{WinoG.} & \textbf{OBQA} & \textbf{PIQA} & \textbf{CSQA} & \textbf{SIQA} & \textbf{SciQ} & \textbf{AVG} \\
\midrule
0.6B & \textbf{45.55} & 26.17 & \textbf{27.74} & \textbf{36.99} & \underline{29.63} & \textbf{52.14} & 28.20 & 65.92 & \underline{20.39} & \underline{38.33} & \underline{64.07} & \underline{39.56} \\
1.7B & \underline{44.75} & \textbf{27.16} & \underline{27.60} & \underline{36.80} & \textbf{30.18} & \underline{51.88} & \underline{29.20} & \underline{66.50} & \textbf{20.56} & \textbf{38.40} & \textbf{66.90} & \textbf{39.99} \\
4B & 44.64 & \underline{26.39} & 27.55 & 36.21 & 29.51 & 51.64 & \textbf{29.73} & \textbf{66.50} & 19.85 & 37.89 & 63.70 & 39.42 \\
\bottomrule
\end{tabular}}
\end{center}
\caption{Effect of the Orchestrator size on downstream performance (0.5B models pretrained from scratch). Each Orchestrator is fine-tuned from the corresponding Qwen3-Base model. The best result in each column is in \textbf{bold} and the second best is \underline{underlined}.}
\label{tab:orchestrator_size}
\end{table}

\subsection{Sampling Parameters at Inference}
\label{app:sampling}

Table~\ref{tab:sampling} lists the decoding parameters used for each model at inference time: the teacher LLM (Qwen3-235B-A22B-Instruct, used for coarse-grained plan generation and verification), the Orchestrator, and the three tool models (NP, SR, PA). For the baseline methods, we instead follow the sampling parameters reported in their original papers.

\begin{table}[h]
\begin{center}
\begin{tabular}{@{}l|ccccc@{}}
\toprule
 & \textbf{Teacher} & \textbf{Orchestrator} & \textbf{NP} & \textbf{SR} & \textbf{PA} \\
\midrule
\texttt{temperature}        & $0.7$  & $0.0$  & $0.0$  & $0.7$  & $0.7$  \\
\texttt{top\_p}             & $0.8$  & $1.0$  & $1.0$  & $0.8$  & $0.8$  \\
\texttt{top\_k}             & $20$   & $20$   & $20$   & $20$   & $20$   \\
\texttt{min\_p}             & $0.0$  & $0.0$  & $0.0$  & $0.0$  & $0.0$  \\
\texttt{presence\_penalty}  & $1.5$  & $0.0$  & $0.0$  & $1.5$  & $1.5$  \\
\texttt{repetition\_penalty}& $1.0$  & $1.0$  & $1.0$  & $1.0$  & $1.0$  \\
\texttt{enable\_thinking}   & False  & False  & False  & False  & False  \\
\texttt{max\_tokens}        & $8192$ & $1024$ & $1024$ & $8192$ & $8192$ \\
\bottomrule
\end{tabular}
\end{center}
\caption{Sampling parameters used at inference time for the teacher LLM, the Orchestrator, and the three tool models (NP, SR, PA).}
\label{tab:sampling}
\end{table}

\section{Details of From-Scratch Pretraining}
\label{app:pretrain}

\paragraph{Models}
For the pretraining experiments in Section~\ref{sec:exp}, we reuse the architectures of Qwen2.5-0.5B, Qwen2.5-1.5B, and Qwen2.5-7B~\citep{qwen2.5}, but randomly initialize all parameters. All three are dense Transformer decoders that use RMSNorm, rotary position embeddings (RoPE), SwiGLU activations, and grouped-query attention. The $0.5$B and $1.5$B models tie their input and output embeddings, while the $7$B model keeps them untied. Table~\ref{tab:pretrain_arch} summarizes the per-size architecture.

\begin{table}[h]
\begin{center}
  \small
  \begin{tabular}{lccc}
    \toprule
    \textbf{Configuration} & \textbf{0.5B} & \textbf{1.5B} & \textbf{7B} \\
    \midrule
    Layers           & $24$        & $28$        & $28$ \\
    Hidden size      & $896$       & $1536$      & $3584$ \\
    FFN hidden size  & $4864$      & $8960$      & $18944$ \\
    Attention heads  & $14$        & $12$        & $28$ \\
    KV groups (GQA)  & $2$         & $2$         & $4$ \\
    Head dimension   & $64$        & $128$       & $128$ \\
    Vocabulary size  & $151{,}936$ & $151{,}936$ & $152{,}064$ \\
    Tied embeddings  & Yes         & Yes         & No \\
    \bottomrule
  \end{tabular}
\end{center}
  \caption{Architectures of the from-scratch pretrained models, following Qwen2.5. All models use RMSNorm ($\epsilon=10^{-6}$), RoPE (base $10^{6}$), and SwiGLU.}
  \label{tab:pretrain_arch}
\end{table}

\paragraph{Training}
We pretrain all models with Megatron-LM~\citep{megatron-lm}. Every run uses a sequence length of $2048$ and a global batch size of $1024$, i.e., about $2.1$M tokens per step. We optimize with Adam ($\beta_1=0.9$, $\beta_2=0.95$, $\epsilon=10^{-8}$), a peak learning rate of $3\times10^{-4}$ held constant after a linear warmup over the first $5\%$ of steps, weight decay $0.1$, and gradient clipping at $1.0$, in bf16. We pack samples to the full sequence length and reset the attention mask and position ids at document boundaries, so tokens never attend across documents. The token budget is $20$B for the $0.5$B and $1.5$B models and $30$B for the $7$B model, corresponding to about $9.5$K and $14.3$K steps. We save a checkpoint every $1000$ steps, i.e., about every $2$B tokens. The remaining hyper-parameters are shared across model sizes and listed in Table~\ref{tab:pretrain_hparams}.

\begin{table}[h]
\begin{center}
  \begin{tabular}{lr}
    \toprule
    \textbf{Hyper-parameter} & \textbf{Value} \\
    \midrule
    Sequence length    & $2048$ \\
    Global batch size  & $1024$ \\
    Optimizer          & Adam ($\beta_1{=}0.9$, $\beta_2{=}0.95$, $\epsilon{=}10^{-8}$) \\
    Peak learning rate & $3\times10^{-4}$ \\
    LR schedule        & constant (after warmup) \\
    Warmup ratio       & $0.05$ \\
    Weight decay       & $0.1$ \\
    Gradient clipping  & $1.0$ \\
    Precision          & bf16 \\
    Training tokens    & $20$B ($0.5$B, $1.5$B) / $30$B ($7$B) \\
    Checkpoint interval & $1000$ steps ($\approx 2$B tokens) \\
    \bottomrule
  \end{tabular}
\end{center}
  \caption{From-scratch pretraining hyper-parameters.}
  \label{tab:pretrain_hparams}
\end{table}

\section{Data Mixture for Math Continued Pretraining}
\label{app:cpt_mixture}

For the math continued-pretraining experiments (Section~\ref{sec:exp}), every method is trained on the same data mixture. The mixture consists of $50\%$ target math data (the math corpus curated by that method), $35\%$ general web data from DCLM-RefinedWeb~\citep{li2024datacomp}, and $15\%$ from the synthetic QA subset of MegaMath~\citep{zhou2025megamath}. The general web data mitigates catastrophic forgetting caused by domain shift, while the synthetic QA data strengthens the model's ability to follow the question-answering format of the downstream generative science-reasoning benchmarks. We use the same ratios for all methods and for both corpora.

\section{Details of Evaluation}
\label{app:eval}

We evaluate all models with our own framework, built on top of the LM-Evaluation-Harness~\citep{eval-harness}. For the general benchmarks, we use likelihood-based (PPL) scoring, which ranks the candidate answers by likelihood instead of generating free-form text. We adopt this format because models pretrained from scratch still have weak instruction-following ability early in training.

The math continued-pretraining experiments (Section~\ref{sec:exp}) additionally use 9 science-reasoning benchmarks: GSM8K~\citep{cobbe2021training}, MATH-500~\citep{hendrycks2021measuring,lightman2024let}, GPQA~\citep{rein2023gpqa}, SuperGPQA~\citep{du2026supergpqa}, SciBench~\citep{wang2023scibench}, the math and STEM splits of MMLU~\citep{hendrycks2020measuring} (MMLU-Math and MMLU-STEM), and the math and STEM splits of MMLU-Pro~\citep{wang2024mmlupro} (MMLU-Pro-Math and MMLU-Pro-STEM). These require free-form answers, so we instead evaluate them in the generative format with few-shot chain-of-thought prompting and greedy decoding, and report exact-match accuracy. Table~\ref{tab:eval_details} summarizes the configuration of all benchmarks.

\begin{table}[h]
\begin{center}
  \small
  \setlength{\tabcolsep}{4.5pt}
  \begin{tabular}{lllccr}
    \toprule
    \textbf{Benchmark} & \textbf{Abbr.} & \textbf{Type} & \textbf{Format} & \textbf{Shots} & \textbf{\# Examples} \\
    \midrule
    ARC-Easy       & ARC-E   & Knowledge             & PPL & $0$ & $2{,}376$ \\
    ARC-Challenge  & ARC-C   & Knowledge             & PPL & $0$ & $1{,}172$ \\
    MMLU           & MMLU    & Knowledge             & PPL & $5$ & $14{,}042$ \\
    HellaSwag      & HellaS. & Language Understanding & PPL & $0$ & $10{,}042$ \\
    RACE           & RACE    & Reading comprehension & PPL & $0$ & $1{,}045$ \\
    WinoGrande     & WinoG.  & Language Understanding & PPL & $0$ & $1{,}267$ \\
    OpenBookQA     & OBQA    & Commonsense reasoning  & PPL & $5$ & $500$ \\
    PIQA           & PIQA    & Commonsense reasoning & PPL & $0$ & $1{,}838$ \\
    CommonsenseQA  & CSQA    & Commonsense reasoning & PPL & $0$ & $1{,}221$ \\
    SIQA           & SIQA    & Commonsense reasoning & PPL & $0$ & $1{,}954$ \\
    SciQ           & SciQ    & Reading comprehension             & PPL & $0$ & $1{,}000$ \\
    \midrule
    GSM8K          & GSM8K      & Science reasoning & Generative & $8$ & $1{,}319$ \\
    MATH-500       & MATH       & Science reasoning & Generative & $4$ & $500$ \\
    GPQA           & GPQA       & Science reasoning & Generative & $5$ & $448$ \\
    SuperGPQA      & SuperGPQA  & Science reasoning & Generative & $5$ & $2{,}582$ \\
    SciBench       & SciBench   & Science reasoning & Generative & $4$ & $688$ \\
    MMLU-Math      & MMLU-M     & Science reasoning & Generative & $4$ & $1{,}064$ \\
    MMLU-STEM      & MMLU-S     & Science reasoning & Generative & $4$ & $2{,}089$ \\
    MMLU-Pro-Math  & MMLU-Pro-M & Science reasoning & Generative & $5$ & $1{,}351$ \\
    MMLU-Pro-STEM  & MMLU-Pro-S & Science reasoning & Generative & $5$ & $4{,}527$ \\
    \bottomrule
  \end{tabular}
\end{center}
  \caption{Per-benchmark evaluation configuration. The two groups correspond to the general benchmarks and the science-reasoning benchmarks used for math continued pretraining, respectively. ``Abbr.'' is the abbreviation; ``Type'' is the capability category; ``Format'' is the evaluation format, either PPL (likelihood-based scoring) or Generative (free-form generation); ``Shots'' is the number of in-context examples; ``\# Examples'' is the size of the evaluation set.}
  \label{tab:eval_details}
\end{table}

\section{Prompt for the End-to-End Rewriting Baseline}
\label{app:e2e_prompt}

The end-to-end variant in Section~\ref{sec:multi_stage} replaces the three NP/SR/PA stages with a single tool LLM that rewrites each chunk in one pass. Its prompt asks the model to triage the chunk and apply noise pruning, surface rectification, and pedagogical augmentation as needed, so that one call covers what \method splits across three specialized stages. The full prompt is shown in Figure~\ref{fig:prompt_e2e}.

\begin{figure}[h!]
\begin{center}
\begin{tikzpicture}
    \node[
        draw=boxstroke,
        line width=1pt,
        rounded corners=10pt,
        fill=boxfill,
        inner sep=5pt,
        align=left,
        text width=0.9\linewidth
    ] (e2eprompt) {
        \textbf{\footnotesize System Prompt}\par\vspace{3pt}
        \begin{Verbatim}[fontsize=\promptfont,breaklines=true,breakindent=0pt,breaksymbolleft={}]
You are an **end-to-end pretraining data cleaner**. Given one chunk of a pretraining document, you produce ONE rewritten version that simultaneously handles three categories of issues -- noise pruning, surface rectification, and pedagogical augmentation -- as needed by this specific chunk. You triage what the chunk actually needs and apply only the relevant operations; not every chunk needs every step.

## Document Format
Each given document chunk is wrapped in [DOC] / [/DOC] tags.
Note: chunks are split by token limits, so the beginning and end may be mid-sentence or mid-paragraph. This is a normal chunking artifact -- do not attempt to complete or extend truncated text at chunk boundaries.

## What to Do

### 1. Noise Pruning
Strip clearly off-content material:
- Website headers, navigation bars, or menu items (e.g., "Home | About | Contact")
- Unrelated HTTP links (e.g., ads, trackers, developer tools)
- Generic footers (e.g., contact info, privacy policies, unsubscribe links)
- Empty lines, decorative separators (e.g., "---"), or stray punctuation runs
For sentences mixing relevant and irrelevant content, remove only the irrelevant fragment if the rest stays coherent; otherwise drop the whole sentence. Never delete meaningful content just because it is short, off-topic-looking at a glance, or stylistically rough -- apply only when content is clearly meaningless.

### 2. Surface Rectification
Fix formatting damage and structural breakage WITHOUT inventing new information:
- Rejoin sentences and paragraphs that were broken by stray line wraps.
- Reconstruct damaged tables, formulas, diagrams, or code blocks that appear as consecutive lines of isolated words, single characters, or short fragments -- use surrounding context to recover the intended structure.
- Normalize obviously broken whitespace, bullets, or list markers.
Limited contextual inference is permitted strictly for restoring the original structure. This is meaning-preserving -- do NOT alter the substance, paraphrase aggressively, or add knowledge that is not already present.

### 3. Pedagogical augmentation
If, after noise removal and surface repair, the content still reads as terse, jargon-heavy, or assumes background a learner may lack, weave in clarifying material:
- Make implicit reasoning explicit (the goal, the strategy, why it fits).
- Demystify jargon and notation with intuition-building analogies and short examples before the formal explanation.
- Illustrate abstract ideas with concrete cases.
- Draw bridges to foundational concepts in the broader field.
- Anticipate likely learner confusion and flag common misconceptions, as natural mentor-style asides.
ALL data, formulas, definitions, theories, experimental results, and logical arguments from the source must be preserved without altering their meaning. Your additions clarify, never contradict. The output must be a self-consistent text that retains elements from the original; do NOT throw away the source and write a fresh explanation from your own knowledge.

## Operating Principles
- Triage first. Decide which of the three categories actually apply to this chunk. A clean, well-written chunk may need only minimal augmentation; a heavily polluted chunk may need aggressive noise removal before anything else.
- Operations compose in order: noise removal -> surface repair -> pedagogical augmentation. Each later stage should treat the cleaned-up text as its input.
- If the chunk after noise removal is empty or trivially short, output the surviving content as-is (or nothing); do not hallucinate a pedagogical narrative from nothing.
- Do not announce what you did. Do not include section headers like "Noise removal:" or "Pedagogical augmentation:". The reader only sees the final rewritten chunk.
        \end{Verbatim}
        \vspace{1pt}\hdashrule{\linewidth}{0.5pt}{2pt}\par\vspace{1pt}
        \textbf{\footnotesize User Prompt}\par\vspace{2pt}
        \begin{Verbatim}[fontsize=\promptfont,breaklines=true,breakindent=0pt,breaksymbolleft={},commandchars=\@\!\?]
[DOC]
@hlchunk!{chunk}?
[/DOC]

## Output Format
output ONLY the final rewritten chunk as a self-contained passage of plain text (no introductory or concluding remarks, no commentary on what was changed, no [DOC]/[/DOC] wrappers).
        \end{Verbatim}
    };
\end{tikzpicture}
\end{center}
\caption{Prompt for the end-to-end rewriting LLM.}
\label{fig:prompt_e2e}
\end{figure}

\section{Prompt for Rewriting-Quality Evaluation}
\label{app:prompt_eval}

In the ablation of Section~\ref{sec:ablation_evolution}, we measure rewriting quality by sampling 700K documents and asking a teacher LLM (Qwen3-235B-A22B-Instruct-2507) to compare each chunk before and after rewriting. Unlike the rewrite verifier used during plan evolution (Figure~\ref{fig:prompt_rewriteverify}), this evaluation only judges two properties of the rewrite, content preservation and factual correctness, and does not write any instruction. To validate the reliability of this evaluation, we conduct a manual review of a sample of the teacher LLM's judgments and find over $86\%$ agreement with human annotations. The prompt is shown in Figure~\ref{fig:prompt_qualityeval}.

\begin{figure}[p]
\begin{center}
\begin{tikzpicture}
    \node[
        draw=boxstroke,
        line width=1pt,
        rounded corners=10pt,
        fill=boxfill,
        inner sep=5pt,
        align=left,
        text width=0.9\linewidth
    ] (qualityeval) {
        \textbf{\footnotesize System Prompt}\par\vspace{3pt}
        \begin{Verbatim}[fontsize=\promptfontsmall,breaklines=true,breakindent=0pt,breaksymbolleft={}]
You are a **rewrite verifier** for pretraining-data cleaning. A cleaning specialist has produced an AFTER version of a document chunk from its BEFORE version -- through format-fixing rewriting, or pedagogical augmentation that unpacks implicit reasoning / demystifies jargon / lowers cognitive load, or both. The specialist must not alter the meaning of the original material. You receive two versions of the SAME chunk and must score the rewrite on a 2-item rubric.

## Input format
- **BEFORE** -- the chunk before the cleaning pass.
- **AFTER**  -- the chunk after the cleaning pass.

Both BEFORE and AFTER are wrapped in [DOC] / [/DOC]. Chunks are token-bounded slices; mid-sentence ends at chunk boundaries are normal.

## Scoring rubric

For each item write `reasoning` first (1-2 sentences), then `score` (the integer `0` or `1` -- `1` means "rubric item passes / no problem").

`no_content_loss` -- Every substantive piece of content in the BEFORE chunk survives in the AFTER chunk. "Substantive" covers facts, numerical values, named entities, definitions, examples, claims, reasoning steps, code, data, etc. Reformatting and interleaving entailed explanatory additions are fine. Removing genuine noisy information is fine and expected -- site navigation, ads, social-share UI, cookie banners, repeated headers/footers, copyright/legal boilerplate, etc. The failure modes to catch: (1) originally-present important information has been deleted, or replaced by a higher-level gloss/summary rather than preserved (e.g. exact numbers, formulas, terminology, quoted phrasing, or code dropped in favour of a paraphrase); (2) the AFTER chunk has been turned into commentary ABOUT the original -- an explanation, description, or meta-discussion of what the BEFORE text says -- rather than remaining a faithful presentation of the text itself, so that the original's own integrity as primary content is lost even if its points are mentioned.

`no_factual_error` -- The rewrite introduced no factually incorrect or hallucinated information. Reconstructing damaged content from context, and inserting reasoning steps, intuitions, analogies, or jargon clarifications, are allowed -- but every such addition must be factually correct and entailed by the BEFORE text (or by uncontroversial domain knowledge naturally implied by it).

## Output format
Respond with EXACTLY one JSON object and nothing else. If you must use a code fence, use ```json ... ```.

```json
{
  "no_content_loss":    {"reasoning": "<sentences>", "score": 0 | 1},
  "no_factual_error":   {"reasoning": "<sentences>", "score": 0 | 1}
}
```

Each `score` is exactly the integer `0` or `1`. Do NOT add any extra keys.
        \end{Verbatim}
        \vspace{1pt}\hdashrule{\linewidth}{0.5pt}{2pt}\par\vspace{1pt}
        \textbf{\footnotesize User Prompt}\par\vspace{2pt}
        \begin{Verbatim}[fontsize=\promptfontsmall,breaklines=true,breakindent=0pt,breaksymbolleft={},commandchars=\@\!\?]
## BEFORE
[DOC]
@hlchunk!{before_chunk}?
[/DOC]

## AFTER
[DOC]
@hlchunk!{after_chunk}?
[/DOC]
        \end{Verbatim}
    };
\end{tikzpicture}
\end{center}
\caption{Prompt for the rewriting-quality evaluation used in the ablation, where a teacher LLM judges content preservation and factual correctness of each rewrite.}
\label{fig:prompt_qualityeval}
\end{figure}

\section{Estimation of Data-Curation Compute}
\label{app:flops}

To compare the cost of different curation pipelines, we estimate the compute each method spends on data cleaning, measured in floating-point operations (FLOPs). Data curation is an inference-time cost: every pipeline runs one or more LLMs over the raw corpus to produce the cleaned data used for pretraining, so we count the inference FLOPs of all LLM passes a method performs during curation. For a Transformer-based language model, the inference FLOPs of a single pass can be approximated as $2N(D_{\text{in}}+D_{\text{out}})$, where $N$ is the (non-embedding) parameter count and $D_{\text{in}}$, $D_{\text{out}}$ are the numbers of input (prefill) and output (decode) tokens~\citep{kaplan2020scaling}. Consider a pipeline of $K$ stages applied in sequence, where stage $k$ uses a model with $N_k$ non-embedding parameters, reads $D_k^{\text{in}}$ tokens, and writes $D_k^{\text{out}}$ tokens. The output of one stage becomes the input of the next. The total curation compute is the sum of the per-stage costs:
\begin{equation}
C_{\text{curate}} \;\approx\; \sum_{k=1}^{K} 2\,N_k \left(D_k^{\text{in}} + D_k^{\text{out}}\right).
\end{equation}
This formulation covers all the methods we compare. For \method we also include the orchestrator's planning and per-chunk decision passes, since they are part of producing the cleaned data. Because curating a pretraining corpus involves processing tens of billions of tokens, the resulting FLOPs are very large. For readability, we report data-curation compute in \emph{exaFLOPs} (EFLOPs), where the prefix ``E'' stands for exa, i.e., $1~\text{EFLOP} = 10^{18}$ FLOPs.

\section{Full Results for Generalization across Datasets and Settings}
\label{app:generalization_full}

Here we report the per-benchmark scores behind Figure~\ref{fig:generalization}. As in the main results, each benchmark score is averaged over the last three checkpoints and AVG is the smoothed average; the best result in each column is in \textbf{bold} and the second best is \underline{underlined}. Tables~\ref{tab:gen_dclm}--\ref{tab:gen_fineweb} cover the three additional web corpora (0.5B, from scratch). Tables~\ref{tab:gen_owm}--\ref{tab:gen_megamath} cover the two math corpora (3B continued pretraining), where the Base row is the model before continued pretraining; MMLU-M/MMLU-S denote the math/STEM splits of MMLU and MMLU-Pro-M/MMLU-Pro-S those of MMLU-Pro.

\begin{table}[t]
\begin{center}
\resizebox{\textwidth}{!}{
\begin{tabular}{@{}lcccccccccccc@{}}
\toprule
\textbf{Method} & \textbf{ARC-E} & \textbf{ARC-C} & \textbf{MMLU} & \textbf{HellaS.} & \textbf{RACE} & \textbf{WinoG.} & \textbf{OBQA} & \textbf{PIQA} & \textbf{CSQA} & \textbf{SIQA} & \textbf{SciQ} & \textbf{AVG} \\
\midrule
Raw & 42.40 & 25.09 & 26.58 & 37.20 & 28.74 & \underline{51.57} & 27.93 & 66.81 & 19.60 & \underline{38.74} & \underline{63.00} & 38.88 \\
Rule-based & 42.31 & 25.48 & 26.66 & 37.83 & \textbf{29.57} & 51.51 & 28.13 & \textbf{67.54} & 19.55 & 38.55 & 62.67 & \underline{39.07} \\
ProX & \underline{43.55} & 25.20 & 27.00 & \underline{38.29} & 28.96 & 50.67 & 26.80 & \underline{67.45} & 19.27 & \textbf{38.96} & \textbf{63.37} & 39.05 \\
RePro & 41.53 & \underline{26.34} & 27.28 & 35.43 & 27.81 & 51.07 & 26.40 & 65.61 & 18.76 & 36.86 & 61.57 & 38.06 \\
ReWire & 42.35 & \textbf{27.67} & \textbf{28.04} & 34.74 & 27.18 & 51.33 & \underline{28.60} & 64.31 & \underline{19.96} & 36.97 & 60.93 & 38.37 \\
\method & \textbf{44.04} & 26.28 & \underline{27.35} & \textbf{38.61} & \underline{29.47} & \textbf{52.25} & \textbf{29.40} & \textbf{67.54} & \textbf{20.04} & 37.82 & 62.47 & \textbf{39.57} \\
\bottomrule
\end{tabular}}
\end{center}
\caption{Per-benchmark downstream performance on DCLM-RefinedWeb (0.5B models pretrained from scratch).}
\label{tab:gen_dclm}
\end{table}

\begin{table}[t]
\begin{center}
\resizebox{\textwidth}{!}{
\begin{tabular}{@{}lcccccccccccc@{}}
\toprule
\textbf{Method} & \textbf{ARC-E} & \textbf{ARC-C} & \textbf{MMLU} & \textbf{HellaS.} & \textbf{RACE} & \textbf{WinoG.} & \textbf{OBQA} & \textbf{PIQA} & \textbf{CSQA} & \textbf{SIQA} & \textbf{SciQ} & \textbf{AVG} \\
\midrule
Raw & 41.44 & 24.72 & 26.79 & 37.82 & 28.42 & 50.78 & 27.87 & \underline{67.65} & 19.57 & \underline{38.30} & \textbf{62.87} & 38.75 \\
Rule-based & 41.30 & 25.34 & 26.76 & \textbf{39.12} & \textbf{29.41} & 51.07 & \underline{28.80} & 67.59 & 19.57 & \textbf{39.15} & 60.30 & \underline{38.95} \\
ProX & 41.20 & 25.80 & 26.70 & \underline{38.97} & 28.74 & 51.01 & 28.73 & 67.65 & 19.57 & 38.08 & 59.57 & 38.73 \\
RePro & 39.93 & 25.65 & 26.76 & 35.56 & 28.42 & 50.88 & 28.73 & 65.05 & \underline{19.96} & 37.46 & 57.50 & 37.81 \\
ReWire & \underline{42.37} & \textbf{28.27} & \textbf{27.36} & 35.26 & 26.44 & \textbf{52.43} & 25.73 & 65.61 & \textbf{20.17} & 36.49 & 60.30 & 38.22 \\
\method & \textbf{44.96} & \underline{26.37} & \underline{27.21} & 38.69 & \underline{28.80} & \underline{51.78} & \textbf{31.80} & \textbf{67.75} & 19.74 & 38.14 & \underline{61.57} & \textbf{39.71} \\
\bottomrule
\end{tabular}}
\end{center}
\caption{Per-benchmark downstream performance on C4 (0.5B models pretrained from scratch).}
\label{tab:gen_c4}
\end{table}

\begin{table}[t]
\begin{center}
\resizebox{\textwidth}{!}{
\begin{tabular}{@{}lcccccccccccc@{}}
\toprule
\textbf{Method} & \textbf{ARC-E} & \textbf{ARC-C} & \textbf{MMLU} & \textbf{HellaS.} & \textbf{RACE} & \textbf{WinoG.} & \textbf{OBQA} & \textbf{PIQA} & \textbf{CSQA} & \textbf{SIQA} & \textbf{SciQ} & \textbf{AVG} \\
\midrule
Raw & 41.89 & 24.57 & 27.03 & 38.05 & \underline{29.60} & 51.20 & 28.40 & 67.12 & 19.57 & \underline{38.81} & \textbf{63.23} & 39.04 \\
Rule-based & 41.84 & 25.20 & 26.70 & \underline{38.48} & 28.23 & 51.91 & 28.33 & 66.96 & 19.57 & \textbf{38.86} & 61.93 & 38.91 \\
ProX & \underline{42.27} & 25.37 & 26.84 & \textbf{38.49} & \textbf{29.70} & \underline{52.30} & \underline{29.67} & \underline{67.65} & 19.49 & 38.23 & 61.33 & \underline{39.21} \\
RePro & 41.44 & 25.03 & 27.22 & 35.39 & 27.88 & 50.17 & 27.00 & 65.49 & 20.01 & 37.65 & 61.33 & 38.06 \\
ReWire & 41.26 & \underline{26.31} & \textbf{27.78} & 35.04 & 27.88 & 51.80 & 26.20 & 64.96 & \textbf{20.64} & 36.27 & 61.83 & 38.18 \\
\method & \textbf{44.46} & \textbf{26.39} & \underline{27.60} & 38.23 & 28.55 & \textbf{53.14} & \textbf{29.87} & \textbf{68.06} & \underline{20.12} & 38.55 & \underline{62.33} & \textbf{39.75} \\
\bottomrule
\end{tabular}}
\end{center}
\caption{Per-benchmark downstream performance on FineWeb (0.5B models pretrained from scratch).}
\label{tab:gen_fineweb}
\end{table}

\begin{table}[t]
\begin{center}
\resizebox{\textwidth}{!}{
\begin{tabular}{@{}lcccccccccc@{}}
\toprule
\textbf{Method} & \textbf{GSM8K} & \textbf{MATH} & \textbf{GPQA} & \textbf{SuperGPQA} & \textbf{SciBench} & \textbf{MMLU-M} & \textbf{MMLU-S} & \textbf{MMLU-Pro-M} & \textbf{MMLU-Pro-S} & \textbf{AVG} \\
\midrule
Base & 19.94 & 11.00 & 9.19 & 8.17 & 3.34 & 30.08 & 26.47 & 10.29 & 10.51 & 14.33 \\
\midrule
Raw & 25.04 & 11.13 & 11.29 & 8.89 & \textbf{3.97} & 35.06 & 25.43 & 10.78 & \underline{12.27} & 15.99 \\
Rule-based & \underline{25.30} & \underline{11.87} & 11.73 & \textbf{9.82} & 3.29 & 33.36 & 25.36 & 10.17 & 11.85 & 15.86 \\
ProX & \textbf{25.90} & 11.13 & \underline{12.63} & 8.47 & \underline{3.92} & \underline{35.12} & 25.95 & 11.35 & 11.74 & \underline{16.25} \\
RePro & 24.64 & 10.73 & 12.18 & \underline{9.44} & 3.00 & \textbf{35.46} & 24.99 & 11.40 & 11.86 & 15.97 \\
ReWire & 24.49 & 10.87 & 12.18 & 8.91 & 3.63 & 34.24 & \underline{27.14} & \underline{11.45} & 12.15 & 16.12 \\
\method & 25.02 & \textbf{12.47} & \textbf{13.23} & 8.53 & 3.83 & \underline{35.12} & \textbf{30.78} & \textbf{12.09} & \textbf{12.38} & \textbf{17.05} \\
\bottomrule
\end{tabular}}
\end{center}
\caption{Per-benchmark science-reasoning performance on OpenWebMath (3B continued pretraining).}
\label{tab:gen_owm}
\end{table}

\begin{table}[t]
\begin{center}
\resizebox{\textwidth}{!}{
\begin{tabular}{@{}lcccccccccc@{}}
\toprule
\textbf{Method} & \textbf{GSM8K} & \textbf{MATH} & \textbf{GPQA} & \textbf{SuperGPQA} & \textbf{SciBench} & \textbf{MMLU-M} & \textbf{MMLU-S} & \textbf{MMLU-Pro-M} & \textbf{MMLU-Pro-S} & \textbf{AVG} \\
\midrule
Base & 19.94 & 11.00 & 9.19 & 8.17 & 3.34 & 30.08 & 26.47 & 10.29 & 10.51 & 14.33 \\
\midrule
Raw & \underline{26.28} & \textbf{12.87} & 12.71 & \underline{9.80} & 3.83 & 35.56 & 25.69 & 11.94 & 11.98 & \underline{16.74} \\
Rule-based & 25.68 & 12.07 & 12.18 & \textbf{10.33} & \textbf{4.12} & 33.49 & 25.34 & 10.81 & 11.62 & 16.18 \\
ProX & 26.21 & \underline{12.13} & \textbf{14.57} & 9.44 & 3.59 & 35.56 & 25.45 & 11.32 & \underline{12.30} & 16.73 \\
RePro & \textbf{26.33} & 11.87 & 11.58 & 9.49 & 3.10 & \underline{36.40} & 25.75 & \underline{11.97} & 12.24 & 16.53 \\
ReWire & 25.78 & 12.07 & 13.08 & 9.48 & \underline{3.88} & 34.12 & \underline{26.76} & 11.70 & 12.08 & 16.55 \\
\method & 25.88 & 12.07 & \underline{13.38} & 8.79 & 3.44 & \textbf{37.22} & \textbf{29.18} & \textbf{12.95} & \textbf{12.56} & \textbf{17.27} \\
\bottomrule
\end{tabular}}
\end{center}
\caption{Per-benchmark science-reasoning performance on MegaMath (3B continued pretraining).}
\label{tab:gen_megamath}
\end{table}

\section{Full Results for Mixtures and Multi-stage Pipelines}
\label{app:extra_full}

Tables~\ref{tab:mix_full} and~\ref{tab:multistage_full} give the per-benchmark scores behind Section~\ref{sec:exp_mixture} (mixtures of rewritten and raw data) and Section~\ref{sec:multi_stage} (multi-stage pipelines), respectively. Each benchmark score is averaged over the last three checkpoints and AVG is the smoothed average.

\begin{table}[t]
\begin{center}
\resizebox{\textwidth}{!}{
\begin{tabular}{@{}lcccccccccccc@{}}
\toprule
\textbf{Method} & \textbf{ARC-E} & \textbf{ARC-C} & \textbf{MMLU} & \textbf{HellaS.} & \textbf{RACE} & \textbf{WinoG.} & \textbf{OBQA} & \textbf{PIQA} & \textbf{CSQA} & \textbf{SIQA} & \textbf{SciQ} & \textbf{AVG} \\
\midrule
Raw & 39.39 & 23.72 & 26.67 & 34.99 & 28.36 & 49.01 & 25.87 & 65.67 & 19.63 & 37.50 & 63.13 & 37.63 \\
ProX & 42.42 & 25.11 & 27.01 & \textbf{37.04} & 29.31 & 52.67 & 26.87 & \textbf{67.03} & 19.55 & \textbf{38.55} & 64.47 & 39.09 \\
RePro & 39.51 & 24.54 & 27.13 & 34.89 & 27.69 & \underline{52.80} & 25.07 & 64.67 & \underline{20.48} & 37.09 & 62.10 & 37.81 \\
RePro + Raw (random) & 40.87 & 24.18 & 26.92 & 35.45 & 28.87 & 50.83 & 27.13 & 65.58 & 19.36 & 36.81 & 63.57 & 38.14 \\
RePro + Raw (fastText) & 44.23 & 24.49 & 27.40 & 35.97 & \underline{29.41} & \textbf{53.27} & 28.80 & 65.58 & 20.34 & 37.97 & 64.37 & 39.26 \\
ReWire & 42.85 & 26.68 & 27.48 & 34.32 & 25.61 & 51.83 & 26.13 & 63.86 & 20.17 & 37.29 & 64.60 & 38.26 \\
ReWire + Raw (random) & 41.92 & 24.29 & 27.17 & 35.47 & 28.52 & 50.64 & 27.20 & 65.49 & 19.52 & 36.56 & 62.23 & 38.09 \\
ReWire + Raw (fastText) & \underline{46.52} & 25.14 & \textbf{28.15} & 34.46 & 28.23 & 50.93 & \underline{29.40} & 64.60 & 20.09 & 37.29 & \textbf{67.33} & 39.29 \\
\method & 44.75 & \underline{27.16} & 27.60 & \underline{36.80} & \textbf{30.18} & 51.88 & 29.20 & 66.50 & \textbf{20.56} & \underline{38.40} & \underline{66.90} & \underline{39.99} \\
\method + Raw (random) & 42.86 & 25.06 & 27.31 & 36.28 & 29.09 & 50.88 & 28.53 & \underline{66.70} & 19.57 & 37.39 & 64.17 & 38.90 \\
\method + Raw (fastText) & \textbf{48.32} & \textbf{27.62} & \underline{28.00} & 35.61 & 29.09 & 52.78 & \textbf{30.33} & 65.80 & 20.45 & 38.26 & 65.23 & \textbf{40.13} \\
\bottomrule
\end{tabular}}
\end{center}
\caption{Per-benchmark results for mixing rewritten data with raw data (0.5B, from scratch). ``(random)'' and ``(fastText)'' denote how the $10$B raw tokens mixed with the $10$B rewritten tokens are selected.}
\label{tab:mix_full}
\end{table}

\begin{table}[t]
\begin{center}
\resizebox{\textwidth}{!}{
\begin{tabular}{@{}lcccccccccccc@{}}
\toprule
\textbf{Method} & \textbf{ARC-E} & \textbf{ARC-C} & \textbf{MMLU} & \textbf{HellaS.} & \textbf{RACE} & \textbf{WinoG.} & \textbf{OBQA} & \textbf{PIQA} & \textbf{CSQA} & \textbf{SIQA} & \textbf{SciQ} & \textbf{AVG} \\
\midrule
Raw & 39.39 & 23.72 & 26.67 & 34.99 & 28.36 & 49.01 & 25.87 & 65.67 & 19.63 & 37.50 & 63.13 & 37.63 \\
Drop & 42.62 & 25.00 & 26.76 & 36.55 & 28.93 & 52.43 & 27.07 & \textbf{67.10} & 19.82 & 37.87 & 63.53 & 38.88 \\
Drop + NP & 42.42 & 25.11 & 27.01 & \textbf{37.04} & 29.31 & \underline{52.67} & 26.87 & \underline{67.03} & 19.55 & 38.55 & 64.47 & 39.09 \\
Drop + NP + SR & 41.54 & 25.71 & 27.09 & 35.31 & 28.68 & 51.51 & 27.67 & 64.87 & 19.71 & 37.31 & 60.97 & 38.22 \\
Drop + NP + SR (fastText) & 44.82 & 26.08 & 27.48 & 36.54 & \underline{29.73} & 51.20 & 28.40 & 65.27 & 19.60 & 37.91 & 64.73 & 39.25 \\
Drop + NP + SR + PA & 45.45 & \textbf{28.58} & 27.77 & 36.19 & 28.26 & 52.07 & \underline{29.20} & 65.02 & \underline{20.67} & \textbf{39.01} & 63.27 & 39.59 \\
Drop + NP + SR + PA (fastText) & \underline{47.12} & \underline{28.01} & 27.82 & 36.52 & 29.25 & 51.14 & 28.80 & 65.03 & \textbf{21.05} & \underline{38.72} & 64.60 & 39.83 \\
Drop + E2E & 43.06 & 25.68 & 27.56 & 35.55 & 28.64 & 51.22 & 28.87 & 65.40 & 20.09 & 37.36 & 60.90 & 38.58 \\
Drop + E2E (fastText) & 46.83 & 26.08 & \textbf{28.28} & 36.68 & 29.60 & 51.67 & 28.13 & 65.56 & 20.31 & 38.52 & 64.70 & 39.67 \\
\method & 44.75 & 27.16 & \underline{28.00} & \underline{36.80} & \textbf{30.18} & 51.88 & \underline{29.20} & 66.50 & 20.56 & 38.40 & \textbf{66.90} & \underline{40.03} \\
\method (fastText) & \textbf{48.32} & 27.62 & \underline{28.00} & 35.61 & 29.09 & \textbf{52.78} & \textbf{30.33} & 65.80 & 20.45 & 38.26 & \underline{65.23} & \textbf{40.13} \\
\bottomrule
\end{tabular}}
\end{center}
\caption{Per-benchmark results for the multi-stage data-processing pipelines (0.5B, from scratch). ``(fastText)'' denotes mixing in fastText-selected raw data.}
\label{tab:multistage_full}
\end{table}

\clearpage
\section{Case Study}
\label{app:case_study}

To give a qualitative sense of the Orchestrator's behavior, we present several real examples from our pipeline. Figures~\ref{fig:case_good_drop}--\ref{fig:case_good_sr} show four \emph{good} cases, and Figures~\ref{fig:case_bad_np}--\ref{fig:case_bad_sr} show two \emph{bad} cases.

\subsection{Good Cases}

\vfill
\begin{casebox}{goodstroke}{goodfill}
{\footnotesize\textbf{Raw Chunk}}\par\vspace{2pt}
\begin{Verbatim}[fontsize=\promptfontsmall,breaklines=true,breakanywhere=true,breakindent=0pt,breaksymbolleft={}]
$19.99 Prime Help! (Mono Vinyl)
$19.99 Prime With The Beatles
$16.12 Prime Help!

B00KZ73VKO,B00KZ73UJQ,B00KZ73W30,B00KZ73V0Y,B0025KVLRY,B0025KVLSS,B00KZ73W08,B00KZ73VVI,B00KZ73VU4,B0025KVLTW,B0025KVLUG,B00KZ73UHI,B00KZ73WCG,B0025KVLV0,B0025KVLTC,B0025KVLVA,B0025KVLU6,B0041KVZ1I,B00GJ7ROX4,B00GJ7ROYS,B00GJ7ROV6,B0000DJZA5,B00GJ7RP10,B00GJ7ROVG,B00GJ7ROXY,B000002USZ,B00O64D3HI,B00GJ7RP4C,B007LS09AU,B00GJ7ROT8,B00NSOP7R6,B000002TYZ,B00GBFC470,B00OCPU6SA,B00GJ7RORU,B00M2OZLQK,B00GJ7ROZW,B00F4BJ4M0,B003YNFYCO,B00M30SPMU,B000002TZ2,B003YNFYD8,B00GJ7RP2O,B00GJ7ROQG,B00M30V1YO,B00IXHBZF6,B00F3VOL38,B00006AW2O,B005NJ9CHK,B00LWLZRIG,B00004ZAV3,B00IXHBS6M,B00M2OZGMY,B00NF4POZO,B00MG4CUH4,B000002TYX,B007L96VCY,B001B0G5WA,B00LRHK3I4,B00KIZ9F8K,B00005NHNX,B00006AW2G,B00NQKWAIQ,B00MI7120K,B003XX2O8C,B00MI711RO,B000RA7IZS,B003Y8YXFS,B00M2OZNBI,B00M5747LY,B00M0CVY3S,B00065XJ48,B000JK8OYU,B0039TD826,B00M574C0U,B00IXHBOF2,B004ZN9T00,B00J2PQYYK,B00AP5KRFI,B000AV2G3I,B00N5NDAB0,B001WCN23W,B00MI711QA,B004WJRF6C,B00EVSACLE,B004E9OLZS,B000VSBX34,B00M2D7KGA,B00BO5SP1Q,B004WDPX9Y,B00NVHRB4W,B00IXHBL7I,B006WTINYI,B00006EXDM,B003Y8YXF8,B00KBUEYMY,B00006AW2J,B00M0CVZ82,B00J4A27B2,B00J2PQZEY
\end{Verbatim}
\vspace{1pt}\hdashrule{\linewidth}{0.5pt}{2pt}\par\vspace{2pt}
{\footnotesize\textbf{Orchestrator Output}}\par\vspace{2pt}
\begin{Verbatim}[fontsize=\promptfontsmall,breaklines=true,breakindent=0pt,breaksymbolleft={}]
{
  "decision": "drop",
  "noise_pruning": false,
  "surface_rectification": null,
  "pedagogical_augmentation": null
}
\end{Verbatim}
\end{casebox}
\captionof{figure}{\textbf{Good case (drop).} The chunk is almost entirely Amazon prices and product identifiers, with no usable linguistic or factual content. The Orchestrator drops it.}
\label{fig:case_good_drop}

\vfill
\begin{casebox}{goodstroke}{goodfill}
{\footnotesize\textbf{Raw Chunk}}\par\vspace{2pt}
\begin{Verbatim}[fontsize=\promptfontsmall,breaklines=true,breakindent=0pt,breaksymbolleft={}]
"Well, that's when we have an overload of hard-boiled eggs," I replied. "But of course that's not really what you're asking."

"No, I guess I'm suggesting that we should make this more often," Mrs K R said. "Fine with me -- we both like it, and it's easy." I chewed thoughtfully. "You know, this new bread recipe you're developing is really good."

"It is," Mrs K R agreed. "My recipe for homemade No-Knead Bread is pretty decent. But I found another method that's even better. And we both like whole wheat bread, so I'm developing this whole wheat version."

"We should put it up on the blog," I said as I finished my sandwich.

"It's not ready for that yet," Mrs K R replied. "We need to test it a bit more -- on egg salad sandwiches!"

Looks like we'll be keeping the Easter Bunny hopping for a while longer this year.
\end{Verbatim}
\vspace{1pt}\hdashrule{\linewidth}{0.5pt}{2pt}\par\vspace{2pt}
{\footnotesize\textbf{Orchestrator Output}}\par\vspace{2pt}
\begin{Verbatim}[fontsize=\promptfontsmall,breaklines=true,breakindent=0pt,breaksymbolleft={}]
{
  "decision": "untouch",
  "noise_pruning": false,
  "surface_rectification": null,
  "pedagogical_augmentation": null
}
\end{Verbatim}
\end{casebox}
\captionof{figure}{\textbf{Good case (untouch).} A fluent, self-contained narrative. Any rewriting would risk distorting the author's voice, so the Orchestrator leaves it unchanged.}
\label{fig:case_good_untouch}

\vfill
\newpage

\begin{centerpage}
\begin{casebox}{goodstroke}{goodfill}
{\footnotesize\textbf{Raw Chunk}}\par\vspace{2pt}
\begin{Verbatim}[fontsize=\promptfontsmall,breaklines=true,breakindent=0pt,breaksymbolleft={}]
a lot of data being sent in this single UDP packet, to a high port and from a high source port. In fact, this is an example of a buffer overflow being completed over UDP. So as we have seen, just using a simple three stage approach we were able to understand what was going on with the scenario. There was an initial port scan for UDP services at 9am. This was then followed by a UDP based buffer overflow attack against 192.168.188 at 10:30am.

David Stubley
7Elements,
We are an independent security consultancy focusing on technical information assurance.
Further information on our services can be found here.
7 Elements to run 44Con Burp Suite Workshop
Stack Exchange Profiles
All data and information provided on this site is for informational purposes only. The opinions expressed by individual Bloggers and those providing comments are theirs alone, and do not reflect the opinions of 7 Elements Ltd. 7 Elements Ltd is not responsible for the accuracy of any of the information supplied by the Bloggers.
(c) 2011, 7 Elements Ltd. Awesome Inc. theme. Powered by Blogger.
\end{Verbatim}
\vspace{1pt}\hdashrule{\linewidth}{0.5pt}{2pt}\par\vspace{2pt}
{\footnotesize\textbf{Orchestrator Output}}\par\vspace{2pt}
\begin{Verbatim}[fontsize=\promptfontsmall,breaklines=true,breakindent=0pt,breaksymbolleft={}]
{
  "decision": "clean",
  "noise_pruning": true,
  "surface_rectification": null,
  "pedagogical_augmentation": "Expand the technical description of UDP-based buffer overflow attacks by adding context about UDP's connectionless nature, its typical use cases (e.g., DNS, RIP), and how such attacks exploit predictable data flows. Ensure all additions are factually grounded in networking principles and do not introduce speculative or unsupported claims."
}
\end{Verbatim}
\vspace{1pt}\hdashrule{\linewidth}{0.5pt}{2pt}\par\vspace{2pt}
{\footnotesize\textbf{After NP} (8 boilerplate lines removed)}\par\vspace{2pt}
\begin{Verbatim}[fontsize=\promptfontsmall,breaklines=true,breakindent=0pt,breaksymbolleft={}]
a lot of data being sent in this single UDP packet, to a high port and from a high source port. In fact, this is an example of a buffer overflow being completed over UDP. So as we have seen, just using a simple three stage approach we were able to understand what was going on with the scenario. There was an initial port scan for UDP services at 9am. This was then followed by a UDP based buffer overflow attack against 192.168.188 at 10:30am.
\end{Verbatim}
\vspace{1pt}\hdashrule{\linewidth}{0.5pt}{2pt}\par\vspace{2pt}
{\footnotesize\textbf{After PA} (cleaned chunk)}\par\vspace{2pt}
\begin{Verbatim}[fontsize=\promptfontsmall,breaklines=true,breakindent=0pt,breaksymbolleft={}]
A lot of data being sent in this single UDP packet -- destined for a high destination port and originating from a high source port -- is a classic sign of a buffer overflow attack carried out over UDP. This might seem surprising at first, because UDP (User Datagram Protocol) is typically seen as a lightweight, connectionless protocol used for real-time applications like DNS queries, streaming media, or routing protocols such as RIP (Routing Information Protocol), where speed and low overhead are prioritized over reliability. Unlike TCP, which establishes a handshake before sending data and ensures orderly delivery, UDP doesn't require a connection setup or flow control. This makes it inherently less secure -- there's no built-in mechanism to detect or reject malformed or oversized packets.

Now, here's the key insight: even though UDP lacks connection establishment, it still operates on predictable patterns. For example, when a device like a router or a DNS server listens on a specific UDP port (say, port 53 for DNS), it will accept incoming datagrams without checking whether they're valid or within expected bounds. Attackers exploit this by crafting oversized packets -- specifically, packets larger than the buffer allocated to handle them -- so that when the receiving service tries to process the data, it writes beyond the end of its memory buffer. This causes a buffer overflow, potentially leading to crashes, unauthorized access, or even remote code execution.

In this scenario, the attack unfolds in two stages. First, at 9 a.m., there's a port scan targeting UDP services -- this means the attacker is probing various high ports to see which ones are open and responsive. The fact that the scan happens early suggests reconnaissance phase, where the attacker identifies potential targets. Then, at 10:30 a.m., the actual attack occurs: a UDP-based buffer overflow is launched against the IP address 192.168.188. The use of a high source port and a large payload in a single UDP packet is designed to overwhelm the target's buffer handling logic.

Why is this dangerous? Because many legacy systems or poorly implemented UDP services don't validate packet sizes before processing. So, even though UDP is fast and efficient, it becomes a vulnerability vector when developers fail to implement proper input validation and boundary checks. This is especially true in network services like DNS, where a simple query can be abused with a crafted oversized response. In contrast, TCP-based attacks often rely on connection states and sequence numbers, making them harder to launch without being detected. But UDP's lack of state tracking and connection management allows attackers to send massive payloads with minimal effort.

So, while this may appear to be a rare or exotic attack, it's actually grounded in real-world vulnerabilities found in older network infrastructure. Understanding this helps us appreciate not just how attacks work, but also why security best practices -- like validating all incoming data, using firewalls to block suspicious traffic, and applying patches to known vulnerable services -- are essential in protecting networks from such exploits.
\end{Verbatim}
\end{casebox}
\captionof{figure}{\textbf{Good case (NP\,+\,PA).} NP removes the consultancy footer and blog boilerplate; PA then expands the terse incident description with background on UDP and buffer-overflow attacks, while preserving the original facts (the 9am scan and the 10:30am attack on 192.168.188).}
\label{fig:case_good_pa}
\end{centerpage}

\newpage
\begin{centerpage}
\begin{casebox}{goodstroke}{goodfill}
{\footnotesize\textbf{Raw Chunk}}\par\vspace{2pt}
\begin{Verbatim}[fontsize=\promptfontsmall,breaklines=true,breakanywhere=true,breakindent=0pt,breaksymbolleft={}]
[DOC]
[000] &gt;&gt;HomeArticlesSign in&nbsp; clickhere
[001] ##Photo synthe sis&nbsp;is how\n plant sconvert</p>
[002] sun lightinto**energy**.\\It happen s&lt;br&gt;in the
[003] chloro plast s wherechloro phyllabsorb slight...|||
...
[010] &nbsp;Related:Bio101 <div>Ads by Google</div>
[/DOC]
\end{Verbatim}
\vspace{1pt}\hdashrule{\linewidth}{0.5pt}{2pt}\par\vspace{2pt}
{\footnotesize\textbf{Orchestrator Output}}\par\vspace{2pt}
\begin{Verbatim}[fontsize=\promptfontsmall,breaklines=true,breakindent=0pt,breaksymbolleft={}]
{
  "decision": "clean",
  "noise_pruning": true,
  "surface_rectification": "Decode and remove residual HTML entities and tags (e.g., &gt;, &nbsp;, <div>, <br>). Merge fragmented tokens that result from faulty text extraction back into complete words. Strip all non-semantic characters, including navigation boilerplate, markdown syntax, escape sequences, and decorative symbols. Reassemble the cleaned content into fluent, well-formed sentences. Do not alter, add, or remove any factual information.",
  "pedagogical_augmentation": null
}
\end{Verbatim}
\vspace{1pt}\hdashrule{\linewidth}{0.5pt}{2pt}\par\vspace{2pt}
{\footnotesize\textbf{After SR} (cleaned chunk)}\par\vspace{2pt}
\begin{Verbatim}[fontsize=\promptfontsmall,breaklines=true,breakindent=0pt,breaksymbolleft={}]
Photosynthesis is how plants convert sunlight into energy. It happens in the chloroplasts, where chlorophyll absorbs light.
\end{Verbatim}
\end{casebox}
\captionof{figure}{\textbf{Good case (NP\,+\,SR).} The extraction is badly corrupted: HTML entities, stray tags, and words split across line breaks. NP finds no clean line-level noise to drop, so SR does the work: it decodes the entities, rejoins the fragmented tokens, and restores two clean sentences without changing the meaning.}
\label{fig:case_good_sr}
\end{centerpage}

\newpage

\subsection{Bad Cases}

\begin{centerpage}
\begin{casebox}{badstroke}{badfill}
{\footnotesize\textbf{Raw Chunk}}\par\vspace{2pt}
\begin{Verbatim}[fontsize=\promptfontsmall,breaklines=true,breakindent=0pt,breaksymbolleft={}]
Provence-Alpes-Cote d'Azur France The Gorges du Verdon, located in south-eastern France (Alpes-de-Haute-Provence), is a river canyon, considered to be one of the most beautiful in Europe. It is about 25 km/16 mi long and up to 700 m/2296 ft deep, formed by the Verdon River and named after for its startling turquoise-green colour. Its most impressive part lies between the towns of Castellane and Moustiers-Sainte-Marie while at the end of the canyon, the Verdon River flows into the artificial lake of Sainte-Croix-du-Verdon. How to go

Wind Surfing ADD COMMENT

Lac de Monteynard, Grenoble
Rhone-Alpes France Lac de Monteynard-Avignonet is an artificial water reservoir serving the Electricite de France power station of Drac Station. The reservoir is bounded by the canyons of Ebron and Drac and belongs to the Isere department. It was created in 1961after the built of a 145 m / 476 ft high dam. The reservoir is up to 10 km / 6.3 mi long and in some places reaches at 300 m / 984 ft wide. On the lake, it is often windy and wavy, something that makes it one of the best places to practice water sports. How to go

Canyoning ADD COMMENT

Durance River, Hautes Alpes
Provence-Alpes-Cote d'Azur France Canyoning in the heart of the Southern Alps, in the Hautes Alpes, on the banks of the Durance river, between the Queyras Regional Natural Park in the east, and the largest National French Park, les Ecrins in the West.

Wakeboarding ADD COMMENT

Rhone-Alpes France Lac de Monteynard-Avignonet is an artificial water reservoir, serving the Electricite de France power station of Drac Station. The reservoir is bounded by the canyons of Ebron and Drac and belongs to the Isere department. It was created in 1961 after the built of a 145 m / 476 ft dam. The reservoir is up to 10 km / 6.3 mi long and in some places it reaches at 300 m / 984 ft wide. On the lake, it is often windy and wavy, something that makes it one of the best places to practice water sports. How to go

Saint Crepin, Hautes Alpes
Provence-Alpes-Cote d'Azur France Saint-Crepin is situated in the Hautes-Alpes department (Provence Alpes region, Cote d'Azur) in the south-east of France at 45 km/27 mi from Gap, the departmental capital. The village is located near the Durance River, ideal spot for whitewater kayaking or canoeing. How to go

Hot Air Ballooning ADD COMMENT

Forcalquier, Haute Provence
Provence-Alpes-Cote d'Azur France Forcalquier is a commune situated in the Alpes-de-Haute-Provence department, in southeastern France. It is located between the Lure and Luberon mountain ranges, about 30 km / 19 mi south of Sisteron and 10 km / 6.2 mi west of the Durance river.

Bungee Jumping ADD COMMENT
\end{Verbatim}
\vspace{1pt}\hdashrule{\linewidth}{0.5pt}{2pt}\par\vspace{2pt}
{\footnotesize\textbf{Orchestrator Output}}\par\vspace{2pt}
\begin{Verbatim}[fontsize=\promptfontsmall,breaklines=true,breakindent=0pt,breaksymbolleft={}]
{
  "decision": "clean",
  "noise_pruning": true,
  "surface_rectification": "Rewrite the document by removing all 'ADD COMMENT' lines and integrating fragmented sentences into coherent paragraphs. Rejoin broken descriptions of geographical locations and ensure all numerical data and proper nouns are preserved exactly.",
  "pedagogical_augmentation": null
}
\end{Verbatim}
\vspace{1pt}\hdashrule{\linewidth}{0.5pt}{2pt}\par\vspace{2pt}
{\footnotesize\textbf{After NP} (all 15 lines removed)}\par\vspace{2pt}
\begin{Verbatim}[fontsize=\promptfontsmall,breaklines=true,breakindent=0pt,breaksymbolleft={}]
(empty -- NP deleted every line, including the geographical descriptions)
\end{Verbatim}
\end{casebox}
\captionof{figure}{\textbf{Bad case (NP over-pruning).} The chunk interleaves genuine geographical descriptions with repeated \texttt{ADD COMMENT} interface lines. NP is meant to strip only the interface noise, but here it deletes every line, including the informative descriptions, leaving an empty chunk. This is an over-pruning failure: NP discards content-bearing text together with the noise.}
\label{fig:case_bad_np}
\end{centerpage}

\newpage

\begin{centerpage}
\begin{casebox}{badstroke}{badfill}
{\footnotesize\textbf{Raw Chunk} (UniProtKB entry for human IMPA2)}\par\vspace{2pt}
\begin{Verbatim}[fontsize=\promptfontsmall,breaklines=true,breakanywhere=true,breakindent=0pt,breaksymbolleft={}]
Searching in BLASTAlignUpload listsContactHelpYou are using a version of browser that may not display all the features of this website. Please consider upgrading your browser.Basket 0(max 400 entries)xYour basket is currently empty.Select item(s) and click on "Add to basket" to create your own collection here (400 entries max)UniProtKB (0)UniRef (0)UniParc (0)AlignBLASTDownloadFull ViewClearO14732- IMPA2_HUMANUniProtO14732 - IMPA2_HUMANProteinInositol monophosphatase 2GeneIMPA2OrganismHomo sapiens (Human)StatusReviewed - <p>Annotation score: 5 out of 5</p>

PublicationsEntry informationMiscellaneousTopBLAST>sp|O14732|IMPA2_HUMAN Inositol monophosphatase 2 OS=Homo sapiens GN=IMPA2 PE=1 SV=1

MKPSGEDQAALAAGPWEECFQAAVQLALRAGQIIRKALTEEKRVSTKTSAADLVTETDHL
VEDLIISELRERFPSHRFIAEEAAASGAKCVLTHSPTWIIDPIDGTCNFVHRFPTVAVSI
GFAVRQELEFGVIYHCTEERLYTGRRGRGAFCNGQRLRVSGETDLSKALVLTEIGPKRDP
ATLKLFLSNMERLLHAKAHGVRVIGSSTLALCHLASGAADAYYQFGLHCWDLAAATVIIR
EAGGIVIDTSGGPLDLMACRVVAASTREMAMLIAQALQTINYGRDDEK

AlignFormatAdd to basketHistoryEntry version 130 (01 Oct 2014)Sequence version 1 (01 Jan 1998)Previous versions | rssHelp videoFeedbackComment Select a section on the left to see content.FunctioniCan use myo-inositol monophosphates, scylloinositol 1,4-diphosphate, glucose-1-phosphate, beta-glycerophosphate, and 2'-AMP as substrates. Has been implicated as the pharmacological target for lithium Li+ action in brain.1 Publication Cited for: FUNCTION, MUTAGENESIS OF ASP-104, ENZYME REGULATION, COFACTOR, BIOPHYSICOCHEMICAL PROPERTIES, HOMODIMERIZATION. Catalytic activityiMyo-inositol phosphate + H2O = myo-inositol + phosphate.CofactoriMagnesium.1 Publication
\end{Verbatim}
\vspace{1pt}\hdashrule{\linewidth}{0.5pt}{2pt}\par\vspace{2pt}
{\footnotesize\textbf{Orchestrator Output}}\par\vspace{2pt}
\begin{Verbatim}[fontsize=\promptfontsmall,breaklines=true,breakindent=0pt,breaksymbolleft={}]
{
  "decision": "clean",
  "noise_pruning": false,
  "surface_rectification": "Reformat the dense, poorly segmented text into clearly separated sections using blank lines. Specifically: (1) separate the 'Catalytic activity' section from the preceding paragraph; (2) ensure that each major section (e.g., 'Cofactor') is visually distinct through spacing.",
  "pedagogical_augmentation": "Expand the 'Cofactor' section by explaining the role of magnesium as the metal cofactor required for inositol monophosphatase activity: describe how Mg2+ ions coordinate the substrate phosphate group and stabilize the transition state during catalysis, and briefly note the enzyme's relevance as the proposed pharmacological target of lithium. Ensure all additions are factually grounded in established enzymology and do not alter, contradict, or remove any existing functional, catalytic, or sequence data."
}
\end{Verbatim}
\vspace{1pt}\hdashrule{\linewidth}{0.5pt}{2pt}\par\vspace{2pt}
{\footnotesize\textbf{After SR}}\par\vspace{2pt}
\begin{Verbatim}[fontsize=\promptfontsmall,breaklines=true,breakindent=0pt,breaksymbolleft={}]
Protein: Inositol monophosphatase 2
Gene: IMPA2
Organism: Homo sapiens (Human)
Status: Reviewed - Annotation score: 5 out of 5

Function:
Can use myo-inositol monophosphates, scylloinositol 1,4-diphosphate, glucose-1-phosphate, beta-glycerophosphate, and 2'-AMP as substrates. Has been implicated as the pharmacological target for lithium Li+ action in brain.

Catalytic activity:
Myo-inositol phosphate + H2O = myo-inositol + phosphate.

Cofactor:
Magnesium.

(the FASTA accession header, the version metadata, and the full 245-residue amino-acid sequence are silently removed)
\end{Verbatim}
\end{casebox}
\captionof{figure}{\textbf{Bad case (SR silent deletion).} The instruction asks SR only to separate sections with blank lines. SR applies the requested formatting but also silently deletes the FASTA accession header, the version metadata, and---most damagingly---the full 245-residue amino-acid sequence, the single most valuable item in a UniProt entry. The output reads more fluently than the input, so no downstream quality filter flags the loss. This exposes SR's bias to treat dense, high-entropy token blocks (sequences, identifiers) as junk and discard them.}
\label{fig:case_bad_sr}
\end{centerpage}

\end{document}